\documentclass[10pt, twocolumn, letterpaper]{article} %
\pdfoutput=1
\usepackage{cvpr}
\usepackage{times}
\usepackage{graphicx}
\usepackage{floatrow}
\usepackage{blindtext}
\usepackage{amsmath,amssymb} %
\usepackage{epsfig}
\usepackage{graphicx}
\usepackage{amsmath}
\usepackage{amssymb}
\usepackage{subfig}
\usepackage{caption}
\usepackage{multirow}

\usepackage{color, bm}

\usepackage[pagebackref=true,breaklinks=true,letterpaper=true,colorlinks,bookmarks=false]{hyperref}
\usepackage{url}

\usepackage[numbers]{natbib}

\DeclareMathOperator{\flow}{FLOW} %
\DeclareMathOperator{\local}{LOCAL} %
\newcommand{\xb}{\mathbf{x}} %
\newcommand{\yb}{\mathbf{u}} %
\newcommand{\zb}{\mathbf{z}} %
\newcommand{\cb}{\mathbf{c}} %
\newcommand{\qb}{\mathbf{q}} %
\newcommand{\Q}{\mathbf{Q}} %
\newcommand{\Qset}{\mathcal{Q}} %
\newcommand{\edges}{E} %
\DeclareCaptionLabelFormat{andtable}{#1~#2  \&  \tablename~\thetable}

\newif\ifkeepFigures
\keepFigurestrue %

\ifcvprfinal\fi

\title{On Pairwise Costs for \\ Network Flow Multi-Object Tracking}
\author{Visesh Chari\thanks{WILLOW project-team, D\'epartment d'Informatique de l'Ecole Normale Sup\'erieure, ENS/INRIA/CNRS
UMR 8548, Paris, France } \qquad Simon Lacoste-Julien\thanks{SIERRA project-team, D\'epartment
d'Informatique de l'Ecole Normale Sup\'erieure, ENS/INRIA/CNRS UMR 8548, Paris, France} \qquad Ivan Laptev\footnotemark[1]
\qquad Josef Sivic\footnotemark[1]\vspace{.3cm}\\
INRIA and Ecole Normale Sup\'{e}rieure, Paris, France\vspace{-.3cm}\\
}

\cvprfinalcopy

\begin{document}
\maketitle

\begin{abstract}

Multi-object tracking has been recently approached with the min-cost network flow optimization techniques.
Such methods simultaneously resolve multiple object tracks in a video and enable modeling of dependencies among tracks. 
Min-cost network flow methods also fit well within the ``tracking-by-detection'' paradigm where object trajectories
are obtained by connecting per-frame outputs of an object detector.
Object detectors, however, often fail due to occlusions and clutter in the video.
To cope with such situations,
  we propose to add pairwise 
  costs to the min-cost network flow framework.
  While integer solutions to such a problem 
  become NP-hard, we design a convex relaxation solution with an efficient rounding heuristic which 
  empirically gives certificates of small suboptimality.
  We evaluate two particular types of pairwise costs 
  and demonstrate improvements over recent tracking methods
  in real-world video sequences.
  \end{abstract}
\vspace{-0.15in}
\section{Introduction}
The task of visual multi-object tracking is to recover spatio-temporal trajectories for a number of objects in a video sequence. Tracking multiple objects, like people or vehicles, has a wide range of applications from Robotics to video surveillance~\cite{yilmaz06objecttracking}. 
Despite recent progress in the field~\cite{andriluka08cvpr,berclaz11pami,butt13cvpr,milan13cvpr,pellegrini09cvpr,pirsiavash11cvpr,yang12cvpr}, tracking remains a challenging problem especially in crowded and cluttered scenes.

\begin{figure}[!t]
\ifkeepFigures
  \centering
  \subfloat[\em No overlap term]{\includegraphics[width=1.58in,height=1.1in]{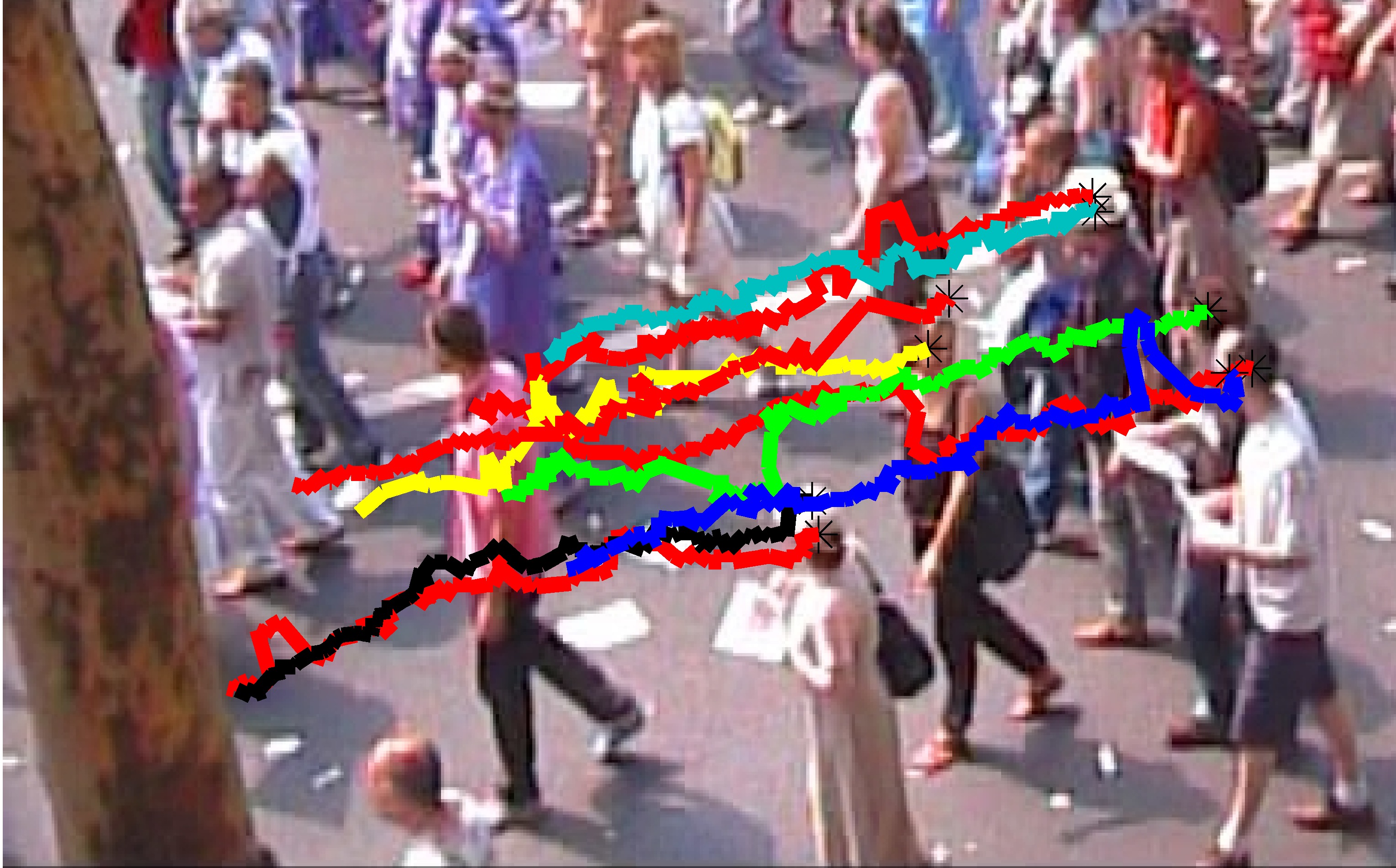}} \,\,      %
  \subfloat[\em With overlap term]{\includegraphics[width=1.58in,height=1.1in]{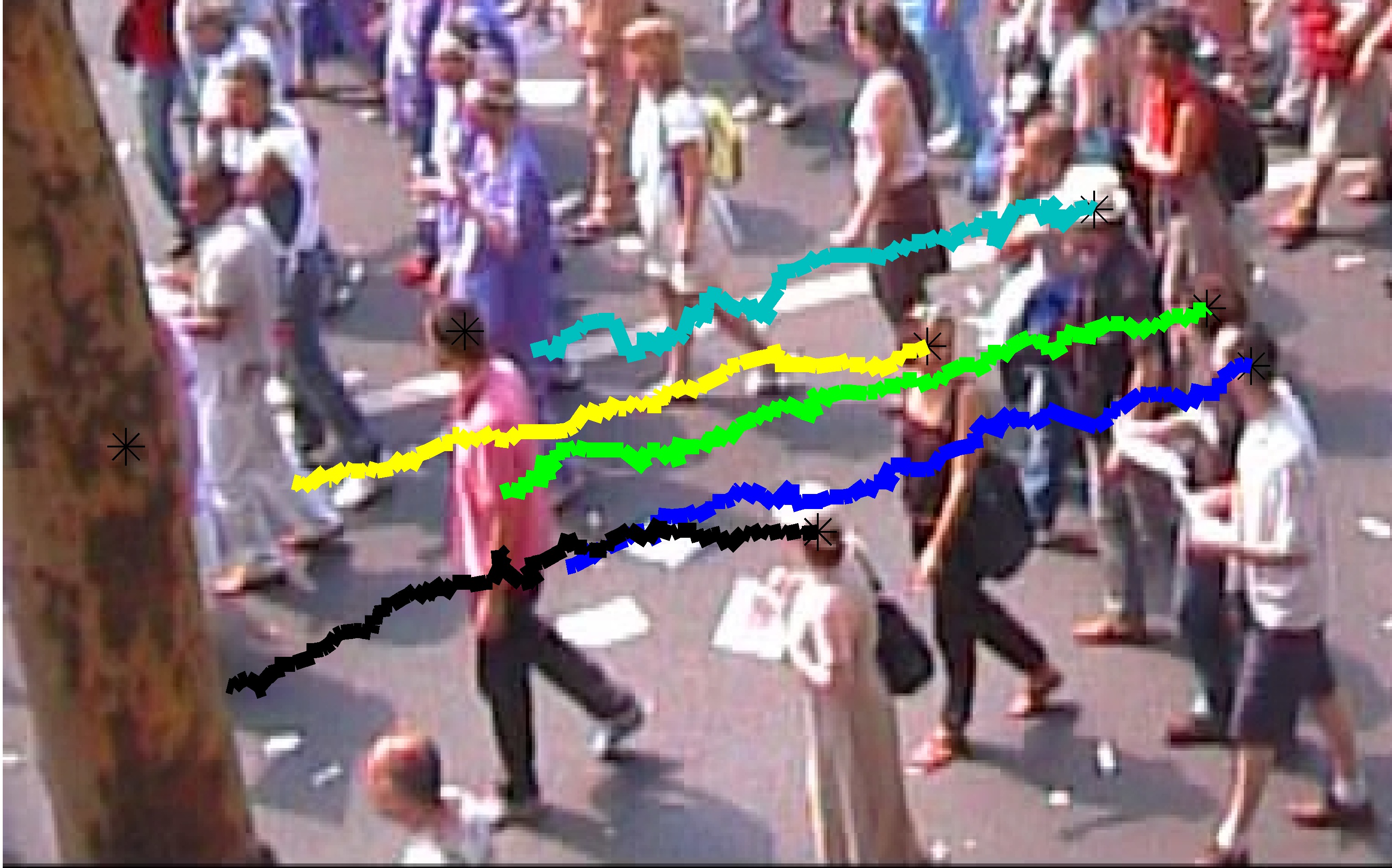}} \vspace{-.2cm}\\    %
  \subfloat[\em No co-occurrence term]{\includegraphics[width=1.58in,height=1.0in]{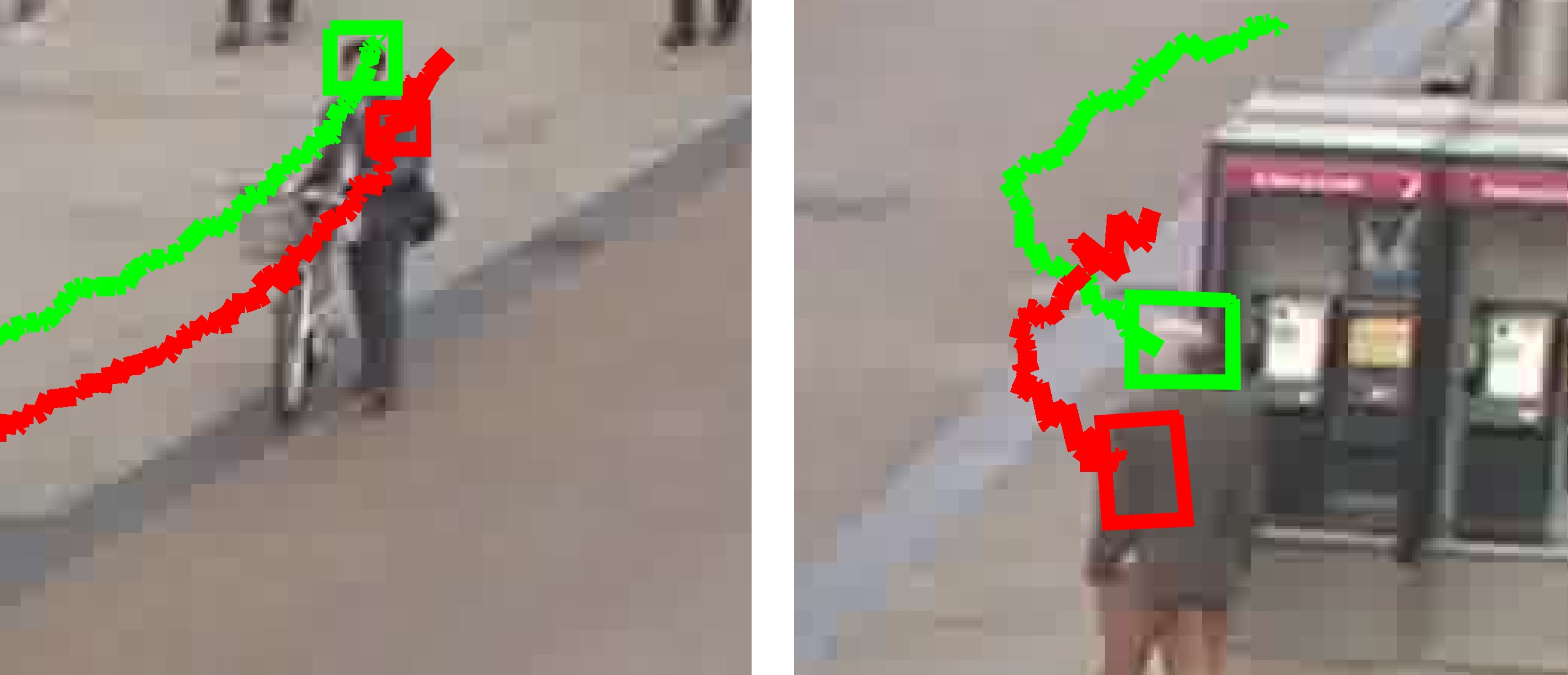}} \,\,   %
  \subfloat[\em With co-occurrence term]{\includegraphics[width=1.58in,height=1.0in]{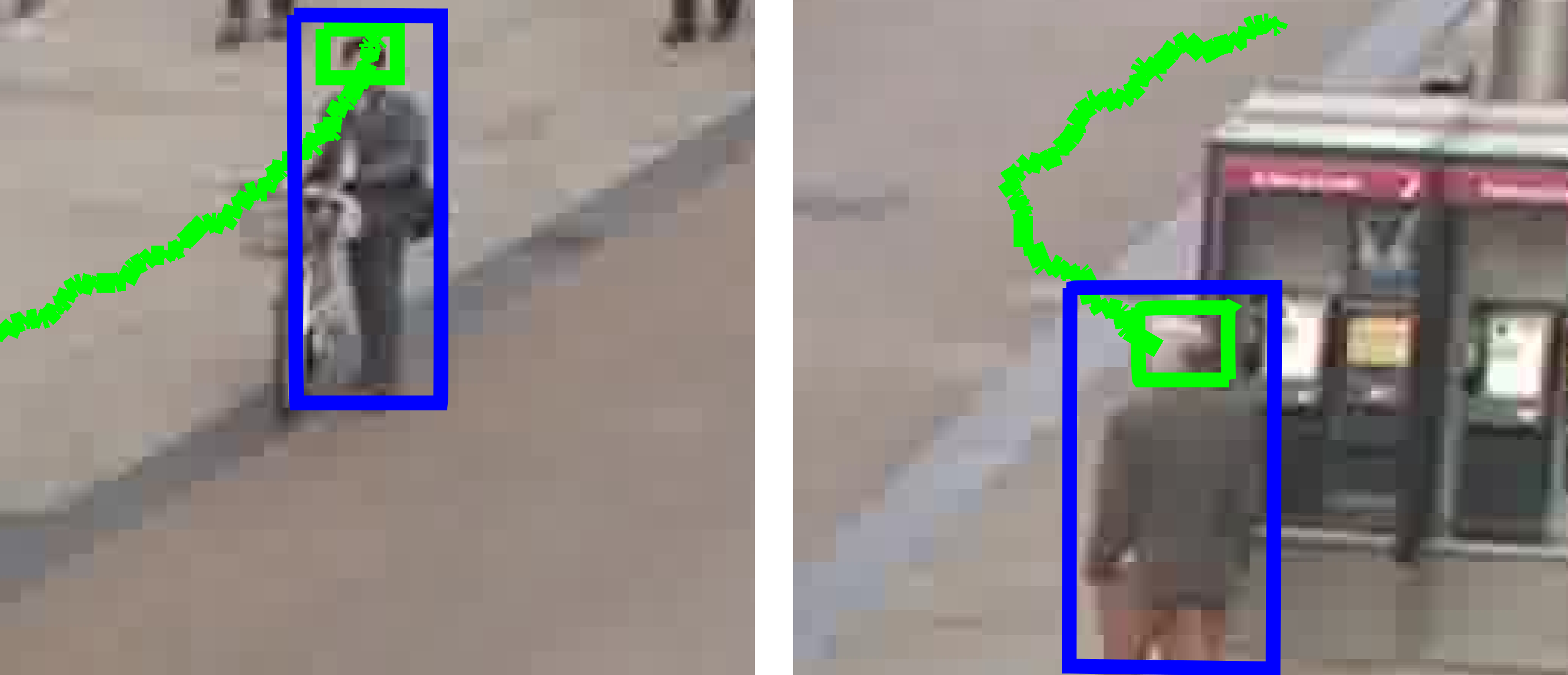}} \vspace{-.1cm}\\ %
\fi
\caption{\small Results of network flow tracking using cost functions with/without pairwise terms.
(a)-(b): a pairwise term that penalizes the overlap between different tracks helps resolving ambiguous tracks (shown in red) in crowded scenes.
(c)-(d): a pairwise term that encourages the consistency between two signals (here head detections and body detections) helps eliminating failures (shown in red) of object detectors.
\vspace{-.5cm}
}
\label{fig:opening}
\end{figure}

With the advances in object detection, ``tracking-by-detection'' have recently become a popular paradigm for object tracking~\cite{berclaz11pami,butt13cvpr,jiang07cvpr,li09cvpr}. 
Given object detections in every frame of a video sequence, the tracking is formulated as selection and clustering of corresponding object detections over time.
Such selection and clustering problems can be solved in an optimization framework using carefully designed cost functions. 
Given an appropriate cost function, tracking-by-detection is typically setup as a MAP estimation problem~\cite{zhang08cvpr}.
Among different formulations of this problem, min-cost network flow~\cite{ahuja93book} is particularly attractive as it allows for optimal and efficient solutions~\cite{pirsiavash11cvpr}. 

The energy minimization approach to tracking enables global solutions to track selection and avoids early and error-prone local decisions.
Moreover, it also enables for a principled modeling of interactions among different tracks.
In the past, models of track interactions have been shown to improve human tracking in crowds~\cite{pellegrini09cvpr}, 
to identify unusual behavior~\cite{kratz09cvpr} as well as to resolve ambiguous tracks~\cite{milan13cvpr,pirsiavash11cvpr}. 
Such previous methods, however, either resort to local {\em non-convex} optimization~\cite{pellegrini09cvpr,kratz09cvpr,milan13cvpr},
or use greedy methods to enforce interactions~\cite{pirsiavash11cvpr}.

Unlike previous work, we here propose to model track interactions within the min-cost network flow tracking approach. 
We introduce pairwise costs to the objective function and design a {\em convex} relaxation solution with an efficient rounding heuristic.
Although our final integer solution can be suboptimal, our method is generic and empirically provides certificates of small suboptimality.
Tracking results using two particular examples of pairwise costs discussed in this paper are illustrated in Figure~\ref{fig:opening}.

In summary, this paper makes the following contributions:
\begin{itemize}
  \item \vspace{-1mm} 
    We propose a new \emph{non-greedy} approach to optimize pairwise terms within 
    a min-cost network flow framework. Our solution is generic and allows the simultaneous 
    optimization of any type of pairwise costs.
	\vspace{-2mm}
  \item We propose a global optimization strategy with a convex relaxation that allows us to minimize
        pairwise costs using linear 
	optimization, and a principled Frank-Wolfe style rounding procedure to obtain integer solutions with a certificate of suboptimality. The optimization procedure is empirically stable, allowing the practitioner to focus on modeling.\hspace{-1mm}
	\vspace*{-2mm} 
  \item %
        To illustrate our method, we propose two particular examples of pairwise costs: the first discourages significant overlaps between distinct tracks; the second models the
        spatial co-occurrence of different types of detections. This allows us	to better model complex dynamic scenes with substantial clutter
        and partial occlusions.
	\vspace{-2mm} 
  \item Using our method, we show improved tracking results on several real-world videos.
        In addition, we propose a new strategy to evaluate tracking results that better measures the longevity 
	of overlap between output tracks and ground truth. 
\end{itemize}
This paper is organized as follows.
Section~\ref{sec:relatedwork} presents related work and the overview of our approach.
Section~\ref{subsec:traditionaltracking} summarizes min-cost flow tracking.
Section~\ref{sec:quadraticcost} describes our optimization
framework with pairwise costs presented in Sections~\ref{subsec:overlap}
and~\ref{subsec:coocurrence}. The optimization strategy is described in
Section~\ref{subsec:optimastrategies}, with initial quadratic optimization formulation in
Section~\ref{sec:quadOpt} and subsequent linear relaxation in
Section~\ref{subsubsec:gradient_based_search}. Finally we present results of our method and compare
them to the state of the art on challenging datasets in Section~\ref{sec:experiments},
and conclude with a discussion in Section~\ref{sec:discussion}.
\section{Related work}

\label{sec:relatedwork}
Recent approaches have formulated multi-frame, multi-object tracking as a min-cost network flow
optimization problem~\cite{zhang08cvpr,pirsiavash11cvpr,berclaz11pami}, where the optimal flow in a connected graph of detections encodes the selected tracks. 
While earlier min-cost network flow optimization methods have used linear programming, recently
proposed solutions to the min-cost flow optimization include push-relabel methods~\cite{zhang08cvpr}, 
successive shortest paths~\cite{pirsiavash11cvpr,berclaz11pami}, and dynamic programming~\cite{pirsiavash11cvpr}. 
To ensure globally optimal and efficient solutions, previous methods have often restricted the cost to unary terms over all edges.
While non-unary terms break the optimality of solutions in general, dependencies between detections have been enforced by
greedy approaches, such as greedily eliminating the overlapping detections after each step of a sequential selection of distinct tracks in~\cite{pirsiavash11cvpr}. This non-global optimization approach, however, cannot recover from early suboptimal decisions.
Additional dependencies among detections can also be incorporated into the min-cost network flow tracking
by modifying the underlying graph structure. Butt and Collins~\cite{butt13cvpr} follows this approach and minimizes the modified objective
using Lagrangian methods. While the method works well for the particular type of introduced cost, 
generalizing this method to the new types of pairwise costs would require appropriate modifications of the graph structure
which is non-trivial in general. Moreover, combining multiple costs within such a framework would be difficult.
In contrast, our framework allows addition of terms without any
modification to the underlying optimization framework.

Brendel et al.~\cite{brendel11cvpr} and Milan et al.~\cite{milan13cvpr,milan13pami} formulate the problem in a framework
that first selects \emph{tracklets} and then connects them using a learned distance
measure~\cite{brendel11cvpr} or a CRF~\cite{milan13cvpr,milan13pami}.
Long term occlusions are handled in~\cite{brendel11cvpr} by merging appearance and motion
similarity. While~\cite{milan13cvpr,milan13pami} propose to alternate between discrete and
continuous optimizations in order to minimize several cost functions, 
the presence of two levels of optimization makes theoretical or empirical guarantees of 
optimality hard to give. Unlike this work, we use a convex relaxation in our approach that allows
us to give an empirical guarantee of optimality to our solutions.

Other methods~\cite{li09cvpr,yang11cvpr,yang12cvpr} use offline or online training to learn a similarity measure between tracklets. 
These methods do not provide any optimality guarantee, though. In addition, training might be difficult in some conditions. For example,
online training to discriminate appearances might be erroneous when objects move
very close to each other (Figure~\ref{fig:opening}). We avoid such problems by using pairwise terms
to robustify the tracker to detection errors.

Incorporation of pairwise terms into the min-cost network flow formulation has been previously attempted by 
Choi and Savarese~\cite{choi12eccv}. Their work, however, is focused on jointly optimizing tracking and activity recognition. 
In contrast, we focus on tracking in particular, and propose a generic framework enabling inclusion 
of multiple types of pairwise costs and providing empirical measures of small suboptimality.

\subsection{Overview of our approach}
\label{subsec:overview}
We propose an algorithm that incorporates quadratic pairwise costs into the traditional
min-cost flow network. Unlike previous methods~\cite{berclaz11pami,li09cvpr}, 
which either build on top of min-cost flow solutions~\cite{milan13cvpr} or change the network
structure~\cite{butt13cvpr}, 
we propose a modification to the standard optimization algorithm.
Such quadratic costs can represent several useful properties like similar motion of
people in a rally, co-occurrence of tracks for different parts of the same object instance and others.

While in such a case obtaining the 
global optimum is NP-hard~\cite{loiola07qap}, we outline an approach to obtain near optimum
solutions, while we empirically verify its optimality. We present a linear relaxation to 
the quadratic term that is fast to optimize, followed by a Frank-Wolfe based rounding heuristic to
obtain an integer solution. 

\section{Background: Min-cost flow tracking}
\label{subsec:traditionaltracking}
In this section, we describe the traditional formulation of multi-object tracking as a min-cost flow optimization problem~\cite{zhang08cvpr}. We extend this framework in Section~\ref{sec:quadraticcost}.

Given a video with objects in motion, the goal is to simultaneously track $K$ moving objects
in a ``detect-and-track'' framework~\cite{zhang08cvpr}. The input to the approach is two-fold.
First a set of candidate object locations is assumed to be given, provided, for example, as output of an object detector.
Henceforth we refer to these locations as {\em detections}. The approach also requires a
measure of correspondence between detections across video frames. This could be obtained for example
from optical flow, or using some other form of correspondence.
Based on these inputs, the tracking problem is setup as a joint optimization problem
of simultaneously selecting detections of objects and connections between them across video frames.
Such a problem can be modeled through a MAP objective~\cite{zhang08cvpr} with specific constraints encoding the
structure of the tracks. The MAP optimization problem can be cast as the following integer linear program (ILP):

\begin{eqnarray}
   \label{eq:mainopt}
 	  \min_{\xb} & \hspace{-2cm}\sum_i c_i x_i + \sum_{ij \in \edges} c_{ij} x_{ij} \\
 	  \mathrm{s.t.} & \left. 
 		\begin{gathered} 0 \leq x_i \leq 1 \, , \, 0 \leq x_{ij} \leq 1  \\
 			\sum_{i \, : \, ij \in \edges} x_{ij} = x_j = \sum_{i \, : \, ji \in \edges} x_{ji}  \\
 			\sum_{i} x_{it} = K = \sum_{i} x_{si} \\
 		\end{gathered} \right\} \xb \in \flow_K \notag \\
 	  & \qquad \quad x_i \, , \, x_{ij} \quad \mathrm{are\quad  integer .} \notag   
\end{eqnarray}

 \begin{figure}[!t]
  \centering
  \includegraphics[width=2.5in,height=1.8in]{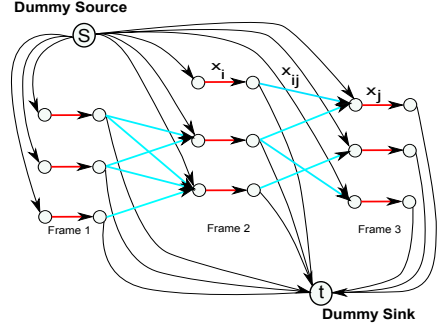}
  \caption{
  Illustration of a graph used in traditional min-cost flow. The detection (red) and
   connection (cyan) variables are marked as edges and have unit capacity. Every track is a unit flow that starts
   at a source S and ends at a sink t. S and t are connected to all detections.
  }
  \label{fig:illustration}
\end{figure}

\noindent
The above formulation encodes the joint selection of $K$ tracks using the following selection variables: 
$x_i \in \{0,1\}$ is a binary indicator variable taking the value $1$ when the \emph{detection} $i$ is selected in some track; $x_{ij} \in \{0,1\}$ is a binary indicator variable taking the value $1$ when detection $i$ and detection $j$ are \emph{connected} through the \emph{same} track in nearby time frames. The index $i$ ranges over possible detections across the whole video. 
$c_i$ denotes the cost of selecting detection $i$ in a specific frame (and represents the negative detection confidence) while $c_{ij}$ represents the negative of the correspondence strength between detections $i$ and $j$. The set of possible connections between detections is represented by $\edges$ and could be a subset of all pairs of detections in nearby frames by using choice heuristics (such as spatial proximity).
The quality of track selection is quantified by the objective in~\eqref{eq:mainopt}. 

The constraint $\sum_{i \, : \, ij \in \edges} x_{ij} = x_j = \sum_{i \, : \, ji \in \edges} x_{ji}$, which has the structure of a \emph{flow conservation constraint}~\cite{ahuja93book}, encodes the correct claimed semantic that $x_{ij}$ can take the value $1$ if and only if both $x_i$ and $x_j$ take the value $1$, and moreover, that each detection belongs to \emph{at most one track}, enforcing the fact that two objects cannot occupy the same space. Finally, the constraint $\sum_{i} x_{it} = K = \sum_{i} x_{si}$ ensures that exactly $K$ tracks are selected (dummy ``source'' and ``sink'' variables with the fixed value $x_s = x_t = K$ are added; the connection variables $x_{si}$ and $x_{it}$ represent the start and end of tracks respectively).

We have grouped the linear constraints in~\eqref{eq:mainopt} under the name $\flow_K$ as they
actually correspond to constraints in a min-cost network flow problem where one would like to push
$K$ units of flow with minimum cost in a network with unit capacity edges. In fact, these linear
constraints have the property of being \emph{totally unimodular}~\cite{ahuja93book}. This implies
that the polytope they determine has only vertices with \emph{integer} coordinates, and so relaxing the integer 
constraints in~\eqref{eq:mainopt} and solving it as a linear program is still guaranteed to produce
\emph{integer solutions}, making it a tight relaxation.
Figure~\ref{fig:illustration} illustrates the correspondence between a network flow structure and the formulation~(\ref{eq:mainopt}).

To summarize, the above optimization problem with relaxed integer constraint can be solved efficiently using existing network flow or linear
algebra packages~\cite{ahuja93book}, and provides
a convenient framework to transform the tracking problem into a \emph{track selection} problem. We use this conversion as a starting point to add additional constraints and costs on the selection process to influence it in desirable
ways to address challenging scenarios that are shown in later sections.

\section{Modeling pairwise costs with an IQP}
\label{sec:quadraticcost}
The above formulation in~\eqref{eq:mainopt} represents a linear objective with linear
equality constraints (where the integer constraint is not needed). 
While linear terms are both simple and easy to minimize, higher order models can represent more 
useful properties~\cite{pellegrini09cvpr}. %
We suggest to add a quadratic cost between pairs of selection variables. 
To simplify the notation for the optimization sections, we collect the $x_i$ and $x_{ij}$ variables
in a long vector $\zb$. The product $z_i z_j$ then encodes \emph{joint} selection of $z_i$ and $z_j$
-- these choices could correspond to a pair of connections, a pair of detections, or even a
connection and a detection. A term of the form $Q_{ij} z_i z_j$ can then either encourage (or discourage) the joint selection of $z_i$ and $z_j$ by having $Q_{ij}$ negative (or positive), respectively. 
Our approach is to consider a small set $\Qset$ of such joint selections, and add the term $\sum_{ij
\in \Qset} z_i z_j Q_{ij}$ to the objective. Our new optimization problem can thus be expressed as the integer quadratic program (IQP):
\vspace{-1mm}
\begin{eqnarray}
\label{eq:quadratic}
\min_{\zb} & \quad \cb^\top \zb + \zb^\top \Q \zb \notag \\
   \mathrm{s.t.} & \quad \zb \in \flow_K \notag \\
   				 & \quad \zb \quad \text{integer ,}
\end{eqnarray}    
where the $\Q$ matrix is \emph{sparse} with $Q_{ij}\ne 0$ for $ij \in \Qset$.

Unfortunately, the above formulation can encode the quadratic assignment problem which is NP-hard to
optimize in general~\cite{loiola07qap}. Nevertheless, we propose an efficient (convex) linear
relaxation in Section~\ref{subsec:optimastrategies} as well as a powerful rounding heuristic that
provides empirical certificates of suboptimality. Our main modeling strategy is thus twofold: first, we encode our prior knowledge about the joint selection of variables using the sparse cost matrix $\Q$ (which can be arbitrary); second we add additional constraints to the IQP as long as they can be encoded as network flow constraints (this is a requirement of our rounding heuristic presented in Section~\ref{subsubsec:gradient_based_search}). In the rest of this section, we provide two examples of pairwise costs used in our experiments. We then focus on the optimization aspects in Section~\ref{subsec:optimastrategies}.

\subsection{Designing pairwise costs}
\label{sec:differentcosts}
In the following subsections, we show how some traditional constraints~\cite{kratz09cvpr,pellegrini09cvpr} 
could be incorporated in our quadratic min-cost network flow framework. We focus on elements that cannot be simply encoded with traditional linear terms in~\eqref{eq:mainopt}.

\subsubsection{Overlap penalty}
\label{subsec:overlap}
Object detectors often produce multiple responses per object. This issue is typically addressed by the \emph{Non Maxima Suppression}~(NMS) step, which retains most confident detections within spatial neighborhoods. While NMS works well for tracking isolated objects, independent decisions produced by NMS for each object and frame often become suboptimal in crowded scenes where multiple objects may occupy the same spatial neighborhood. 
To address this problem, we avoid taking independent decisions and propose to discourage overlapping detections within the network flow tracking framework. For this purpose we extend the cost function with the following {\em pairwise overlap cost}:
\begin{gather}
\label{eq:overlapterm}
  q_{ij}^{\mathrm{ov}} x_i x_j \\
  \textrm{for} \,\,\, (i,j) \,\,\, \textrm{s.t.} \,\,\, \mathrm{ov}( \mathrm{box}(x_i), \mathrm{box}(x_j) ) \geq o_{\mathrm{thres}}  \nonumber
\end{gather}
where $x_i$ and $x_j$ represent two selection variables associated with sufficiently overlapping\footnote{The overlap threshold $o_{\mathrm{thres}}$ is set to 0.5 in our experiments.} detections and $q_{ij}^{\mathrm{ov}} > 0$. 

In previous approaches like~\cite{pirsiavash11cvpr}, NMS was implemented in a \emph{greedy} fashion.
Greedy approaches, however, have the disadvantage of making non-reversible decisions in the early stages of optimization.
In contrast, our approach of incorporating the cost~(\ref{eq:overlapterm}) into the overall cost function ensures that
NMS is optimized \emph{simultaneously} with other tracking objectives. As a result, overlapping detections
may become tolerated, for example, in situations when two tracks intersect. On the other hand, continuously
overlapping tracks resulting from multiple outputs of  detectors will be discouraged.

\subsubsection{Enforcing consistency between two signals}
\label{subsec:coocurrence}
In many tracking scenarios, multiple signals are available for use. For example, we might have a body detector as well as a head detector. In case they give
complementary information about the presence of the object, we can be more robust to detection noise by ensuring that the two tracks are consistent using a pairwise cost.

For example, let $z^h_i$  and $z^b_i$ denote the selection variables (detection or connection) for the head and body respectively. Each set can be associated with its own flow feasible
set $\flow^h_K$ and $\flow^b_K$. 
We can \emph{encourage} the consistent ``co-occurrence'' of the two flows by adding the following negative cost:
\vspace{-3mm}
\begin{gather}
  \label{eq:coocurrence}
  -q_{ij}^{\mathrm{co}} z_i^h z_j^b \\
  \textrm{for $(i,j)$ s.t. $z_i^h$ and $z_j^b$ are \emph{consistent}}. \nonumber
\end{gather}
In our experiment, we say that $z_i^h$ and $z_j^b$ are \emph{consistent} in two scenarios. Either~$z_i^h$ and $z_j^b$ are detection variables such that their corresponding boxes\footnote{For the body detection box, we only consider its top 25\% region when computing overlap or looking at intersection.} overlap more than $o_{\mathrm{thres}}$. 
Or we have a head detection $z_i^h$ with a box that intersects the edge $z_j^b$ connecting its respective body detection boxes (and similarly for a body detection and head edge). The idea behind the latter possibility is to be more robust to missing detections on some frames: it corresponds to a situation where a head and body detection would have overlapped if we were interpolating detections along an edge that skips frames. Note that the cost~\eqref{eq:coocurrence} is difficult to minimize greedily, since both head and body tracks need to be optimized \emph{simultaneously}.
\section{Optimization}
\label{subsec:optimastrategies}
In the previous section, we presented examples of quadratic cost functions that we could include in our extension to the min-cost flow network formulation to encourage co-occurrence preferences for individual variables in the minimization. Finding a global minimum is NP-hard~\cite{loiola07qap} if we keep the integer constraints on the variables (which is necessary to ensure the correct track encoding). Our suggested strategy is to instead find a global solution to the \emph{relaxed} version
of the problem with the integer constraints removed, and then use a powerful heuristic to search for nearby integer solution that satisfies the flow constraints (see Section~\ref{subsubsec:gradient_based_search}).  
By comparing the objective value between the ``rounded'' integer solution and the global solution to the relaxed problem, which provides a lower bound, we obtain a \emph{certificate of optimality}. In our experiments, we observed that suboptimality upper bounds were quite small, thus indicating that our optimization framework is stable and we can instead focus on designing good cost functions. 
We now describe several approaches to optimize~\eqref{eq:quadratic}.
\subsection{Quadratic optimization} \label{sec:quadOpt}
If $\Q$ is positive definite, then the quadratic program (QP) in~\eqref{eq:quadratic}
with relaxed integer constraint is convex and can be robustly optimized using
interior point methods implemented in commercial solvers such as MOSEK/CPLEX. These
methods can scale to medium-size problems\footnote{A few millions variables, which translates to
several hundreds frames with a high number of detections for our datasets.} by exploiting the
sparseness of $\Q$ suggested in Section~\ref{sec:differentcosts}. 

In our general formulation, $\Q$ is not necessarily positive definite. 
We can nevertheless use a standard trick to make it positive definite by defining its diagonal entries to 
be $Q_{ii}^{\mathrm{new}} = \sum_{j\neq i} |Q_{ij}^{\mathrm{old}}|$, while using $c_i-Q_{ii}^{\mathrm{new}}+Q_{ii}^{\mathrm{old}}$ 
as the linear coefficient for $z_i$ in the objective. As $z_i^2 = z_i$ for binary variables, this
transformation sill yields an (equivalent) IQP. %
$\Q^{\mathrm{new}}$ is now positive semidefinite~\cite[Thm. 6.1.10]{hornAndJohnson}, and so the relaxation gives a convex problem.

In order to scale to very large scale datasets (billions of variables), one could use the Frank-Wolfe
algorithm~\cite{jaggi13icml} which is a first order gradient based method that iteratively minimizes
a linearization of the quadratic objective. An advantage of this approach is that each step of the
Frank-Wolfe algorithm reduces in our case to the minimization of a min-cost network flow problem,
which can scale to much larger sizes than a generic linear program solver. Moreover, each step of
this algorithm yields an integer solution. Thus, while optimizing the relaxed objective (which will
provide a lower bound certificate), we can keep track of which integer iterate had the best
objective thus far. This perspective also motivates a powerful rounding heuristic that we describe in Section~\ref{subsubsec:gradient_based_search}. Building on a preliminary version of our paper, \cite{joulin14FW} used this approach successfully for performing efficient co-localization in videos, where the constraint set also had a network flow structure.

\subsection{Equivalent integer linear program}
\label{subsec:linearconstraints}
Another way to make the approach more scalable is to transform the integer QP~\eqref{eq:quadratic} into an equivalent integer linear program (ILP) by introducing well-chosen additional variables and constraints. We present such an approach in this section, which generalizes the line of reasoning from~\cite{lacoste06qap}.

We introduce a new set of variables $u_{ij}$ that encode the joint selection of the edge $z_i$ and
$z_j$, and thus we would like to enforce $u_{ij} = z_i z_j$. The quadratic cost component $Q_{ij}
z_i z_j$ could then be replaced with a linear cost $Q_{ij} u_{ij}$.
An equivalent integer linear
program is thus the following:
\begin{eqnarray}
\label{eq:quadlinear}
   \min_{\zb, \yb} & \qquad \quad \cb^\top \zb + \qb^\top \yb  \notag \\
    & \qquad \quad \zb \in \flow_K \notag \\
   \mathrm{s.t.} & \quad \left.  
   	\begin{gathered}  0 \leq u_{ij} \leq 1, \, \forall ij \in \Qset \notag \\
   		  u_{ij} \leq z_i \, , \, u_{ij} \leq z_j \notag \\     			
   		  z_i + z_j \leq 1 + u_{ij} \,  \\
   	\end{gathered} \right\} 
   	   	\begin{gathered}
   	   	(\zb, \yb) \in \\
   	   	  \local(\Qset)\\
   	   	\end{gathered} \\
  &  \qquad \quad \zb, \yb \quad \text{integer .}  
\end{eqnarray}

Here $\yb$ and $\qb$ represents the vector whose elements are $u_{ij}$ and $Q_{ij}$ respectively. 
The new constraint $ z_i + z_j \leq 1 + u_{ij}$ enforces that $u_{ij}$ should be $1$ if $z_i$ and
$z_j$ are both $1$; while the pair of constraints $u_{ij} \leq z_i$ and $u_{ij} \leq z_i$ enforce
$u_{ij} = 0$ if either $z_i$ or $z_j$ is zero. We call these constraints
`$\local(\Qset)$' as it turns out that they define a polytope which can be obtained by a projection
of the \emph{local marginal consistency} polytope for the over-complete representation of a discrete
Markov random field (MRF)~\cite[(4.6)]{wainwright08} with edges defined by the non-zero entries of
$\Q$\footnote{More specifically, this representation defines one indicator variable per possible
joint assignment of values on the cliques of the MRF. If we do Fourier-Motzkin
elimination~\cite{bertsimas,wainwright08} on the local consistency polytope to eliminate the extra
variables and to only keep the three variables $z_i, z_j, u_{ij}$ for each edge, then we obtain back
the constraints for $\local(\Qset)$.}. Removing the integer constraint in~\eqref{eq:quadlinear} thus
yields a LP relaxation that is similar to one for MAP inference in MRFs, but with additional
$\flow_K$ constraints, yielding a crucial structural difference with the previous works.

An advantage of this formulation is that its relaxed form is a LP, which can usually be optimized by MOSEK or CPLEX to larger scale than the QP formulation, even though there is an increase in the number of variables and constraints.
Note though that the number of new variables $u_{ij}$ created is the same as the number of non-zero coefficients in the sparse $\Q$, which was indicated by the set $\Qset$ in~\eqref{eq:quadlinear} to stress that we do not need to look at all pairs of edges.
In exploratory experiments, we observed that the LP relaxation yielded similar
quality solutions as the QP relaxation, but was faster to optimize; we have thus focused on the LP
relaxation in our experiments. Another advantage of~\eqref{eq:quadlinear} is that we
can easily generalize it to handle higher order terms in the objective. For a clique $C$ of decision
variables that we want to encourage or discourage jointly, we introduce a new variable $u_C :=
\prod_{i \in C} z_i$. This semantic can be readily enforced with the constraints $u_C \leq z_i$ for
all $i \in C$, and $\sum_{i \in C} (z_i-1) + 1 \leq u_C$, which generalizes $\local(\Qset)$ for
higher order terms and yields another ILP that can be relaxed to a LP.

\subsection{Frank-Wolfe rounding heuristic}
\label{subsubsec:gradient_based_search}
The solution of the LP relaxation of~\eqref{eq:quadlinear} can have fractional components because the  additional linear constraints from $\local(\Qset)$ essentially violate the \emph{total unimodularity} property, in contrast to $\flow_K$ which yields a polytope with only integer vertices. Since naively rounding the obtained fractional variables to the nearest integer might not result in a feasible point
(in other words a valid flow), we need a strategy to obtain an integer solution with cost similar to
the minimum. Given the relaxed global solution $\zb^*$, the simplest approach would be to look for the point closest in Euclidean norm in $\flow_k$ which is an integer. As $z_i^2 = z_i$ for binary variables, we have $||\zb-\zb^*||^2 = (\bm{1}-2\zb^*)^\top \zb + ||\zb^*||^2$ which is a linear function of $\zb$. We can thus obtain the closest integer point by solving a LP over $\flow_k$, as all its vertices are integers. We call this approach \emph{Hamming rounding} as $d_H(\zb,\zb'):=||\zb-\zb'||^2$ reduces to the Hamming distance when evaluated on pair of binary vectors. On the other hand, the closest point in Euclidean norm does not necessarily yield a good objective value (as the search was agnostic to the objective). Inspired by the Frank-Wolfe algorithm, our suggested heuristic is to minimize instead the first-order linear under-estimator of the quadratic objective constructed with the gradient at the relaxed global solution $\zb^*$. 
Specifically, we obtain the following LP, which has the usual network flow constraint structure and thus can be solved very efficiently:
\begin{eqnarray}
\label{eq:gradsearch}
\min_{\zb} & \quad \left( \cb + (\Q + \Q^{\top}) \zb^*\right)^\top \zb \notag \\
   \mathrm{s.t.} & \quad \zb \in \flow_K .
\end{eqnarray}
The objective here can be interpreted as modifying the distance function on binary vectors to take
the cost function in consideration. As previously mentioned, the relaxed LP solution provides a
lower bound on the true ILP (which is equivalent to the IQP) solution. The difference between the
objective evaluated on \emph{any} feasible integer solution and the lower bound is thus an {upper
bound certificate} on its suboptimality. 
In our experiments, we obtained small suboptimality
certificates ($\approx 10^{-3}$) for our returned integer solutions, indicating that our rounding
heuristic was effective at returning near-global optimal solutions (we note that we define $\cb$ and $\Q$ so that the objective is normalized between $0$ and $1$). 
We also observed that Hamming rounding generally produced a suboptimality that was around $3$ to $4$ times \emph{worse} than the solution produced by Frank-Wolfe rounding. These worse objective values also translated in worse tracking accuracy (see Appendix~\ref{sec:FWbetter} in the supplementary material\footnote{The supplementary material (with videos and code) is available at~\cite{projectwebpage}.}).
We finally note that in contrast to the previous work~\cite{butt13cvpr} which could not guarantee that their algorithm would converge to an integer solution, our approach will always give \emph{some} integer solution (by solving a simple min-cost network flow problem), and can provide a certificate of suboptimality a-posteriori.

\section{Experiments}
\label{sec:experiments}
In this section, we evaluate our approach on several real world videos and compare results to
the state-of-the-art methods~\cite{benfold2011stable,milan13cvpr,pirsiavash11cvpr}. First we illustrate the effect of 
the two pairwise costs proposed in Section~\ref{sec:differentcosts} and evaluate improvement over
the basic min-cost network flow tracking. We also argue that the standard MOTA score is often 
insufficient to capture the quality of tracking results and propose
a new measure for tracking evaluation, termed \emph{re-detection measure}~(Section~\ref{para:redetection}).

Second, we evaluate our method on six videos from the two standard datasets PETS and TUD. For both of these 
datasets, we obtain part of the input (person detections) from Milan et
al.~\cite{milan13cvpr}, and show improvements over their approach using the standard MOTA metrics.

\subsection{Tracking datasets}
We test our algorithm on several publicly available videos. %
The first video MarchingRally corresponds to a crowd walking in a rally along a street (see Figure~\ref{fig:opening}, top row).
The video consists of 120 frames recorded at 25 fps, and has about 50 people. This video is challenging due to the high number of people moving
close to each other. We have manually annotated ground truth tracks for all people in this video for the purpose of tracking 
evaluation\footnote{The original MarchingRally video and the corresponding ground truth tracks are available from~\cite{projectwebpage}.}.

The second video illustrated in Figure~\ref{fig:opening} (bottom row) is called
TownCenter~\cite{benfold2011stable}
and consists of 4500 frames recorded at 25 fps. The video shows approximately 230 people 
walking across the street.
Finally, we use videos from the well-known PETS and TUD datasets. These videos depict frequently occluded 
people moving in multiple directions.

\vspace{-3mm}
\paragraph{Preprocessing.}
We run a ``head'' detector~\cite{mikel11iccv} to detect heads of people in every frame of 
the MarchingRally and PETS videos. While we use only head detections for the MarchingRally
sequence, for PETS we use our head detections in combination with readily-available body detections from~\cite{milan13cvpr}.
Head detections complement 
frequently overlapping body detections and help resolving partial occlusions as
well as ID-switches.
For each of these videos,
we run a KLT tracker after initializing features within detection bounding boxes.
Finally, for every pair of nearby
frames ($<$~10 frames apart), we connect pairs of detections with
high correspondence strength.
The strength of correspondence between two detections is the ratio of their common
KLT tracks and the total number of KLT tracks passing through both detections.

\subsection{Tracking in video experiment} \label{sec:trackingExp}
\subsubsection{Evaluation strategy}
\label{subsec:strategy}
Evaluating results of multi-object tracking is non-trivial because errors might be present in
various forms including ID switches, broken tracks, imprecisely localized tracks and false tracks.
Measures such as MOTA~\cite{benfold2011stable,milan13cvpr} combine different errors into
a single score and enable the global ranking of tracking methods. Such measures, however, lack interpretability.
On the other hand, independent assessment of different errors can also be misleading.
For example, in dense crowd videos such as in Figure~\ref{fig:opening}(a), tracks may have relatively low
localization error while being incorrect due to switches between neighboring people. 
Similarly, low error of ID switches can be a consequence of many broken tracks. 

We argue that a meaningful evaluation of tracking methods should be related to a task.
One task with particular relevance to crowd videos is to detect the location of a given person after 
$\Delta t$ frames. To evaluate the performance of tracking methods on such a 
task we propose the re-detection measure as described below.

\vspace{-2mm}
\paragraph{Re-detection measure.}
\label{para:redetection}
The proposed re-detection measure evaluates the ability of a tracker to find the correct location of a given object
after $\Delta t$ frames. The measure is inspired by the common evaluation procedure for object detection in still 
images~\cite{everingham10} and extends it to tracking. For each pair of detections $A_t$ and $B_{t+\Delta t}$ associated to 
the same track by a tracker, we check if there exists a ground truth track that overlaps with $A_t$ and $B_{t+\Delta t}$ on
frames $t$ and $t+\Delta t$ respectively.\footnote{The overlap between ground truth and detections is measured by the standard 
Jaccard similarity of corresponding bounding boxes.} If the answer is negative, the subtrack $(A_t,B_{t+\Delta t})$ is 
labeled as false positive. Otherwise, it is labeled as true positive unless there exist multiple subtracks overlapping
with the same ground truth. To avoid multiple responses, in the latter case only one subtrack
is labeled as true positive while others are declared as false positives.

\begin{figure}[t]
\begin{minipage}{.80\textwidth}
\ifkeepFigures
  \subfloat[]{\includegraphics[width=1.3in,height=1.3in]{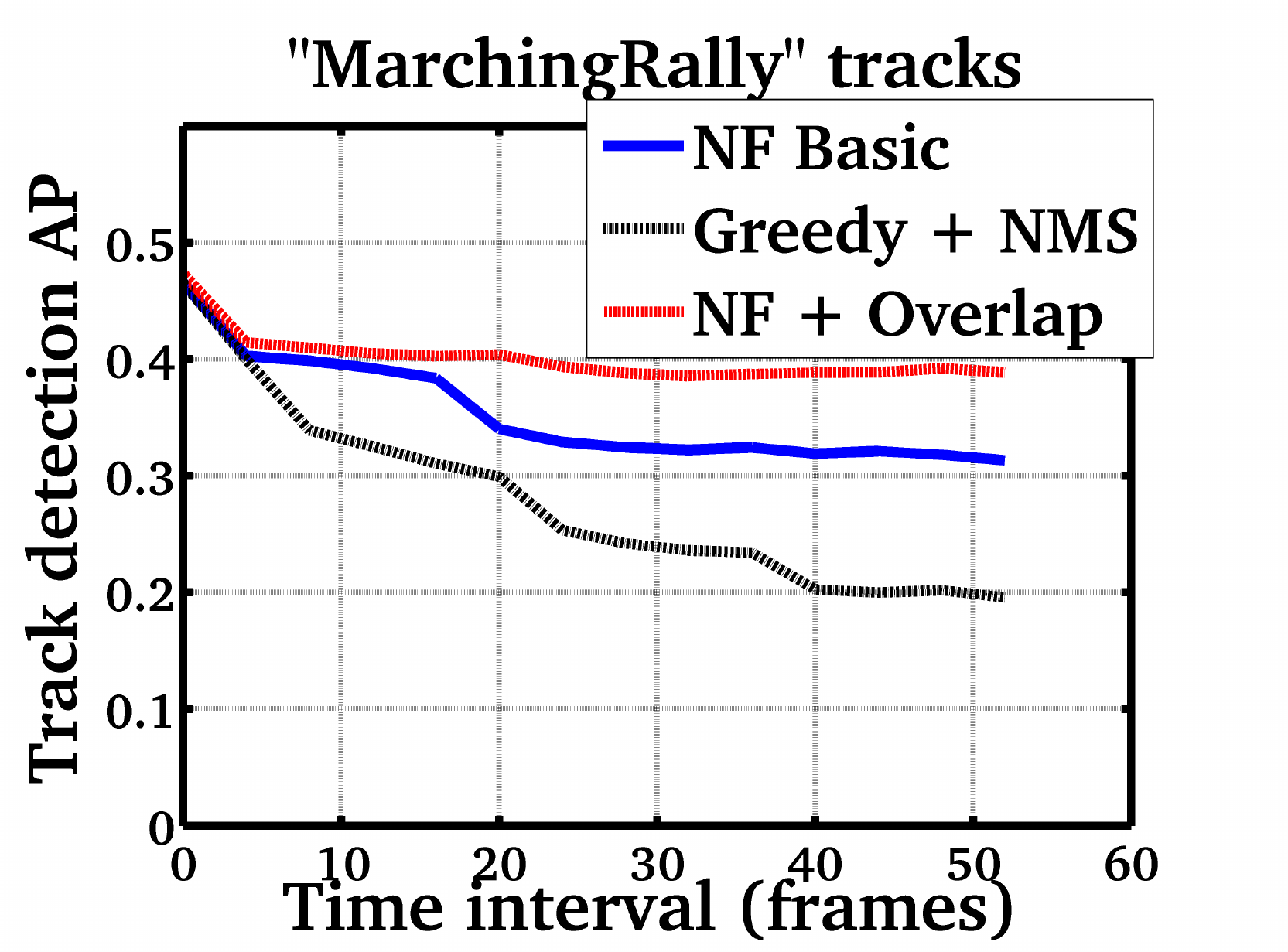}} \hfill
  \subfloat[]{\includegraphics[width=1.3in,height=1.3in]{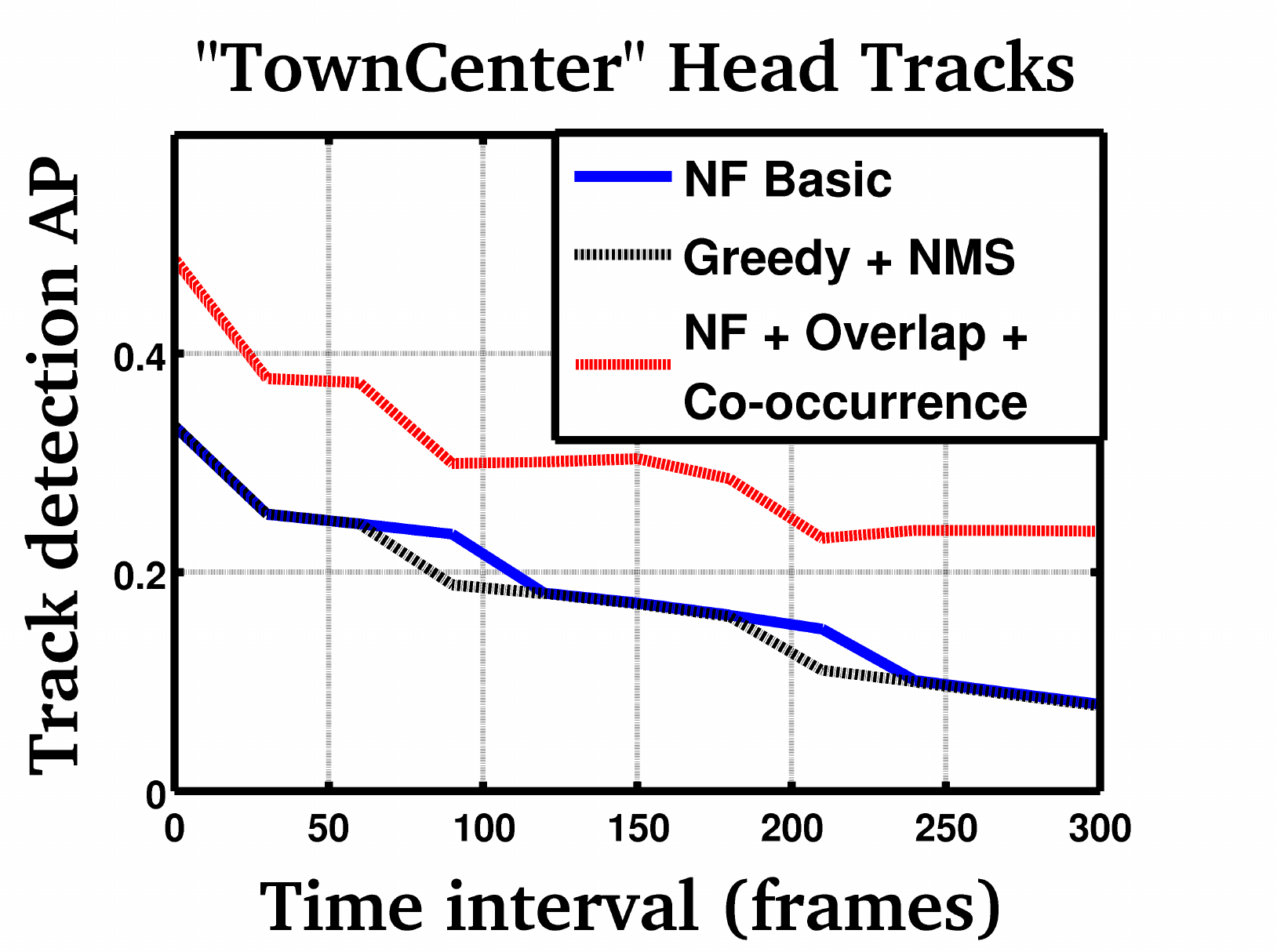}} 
\fi
\end{minipage} \\
\begin{minipage}{.80\textwidth}
  \small
  \begin {tabular}{ |c||c|c|c|c| }
  \hline  & MOTA & Prec & Recall \\ \hline
  NF + Ov. + Co-oc. & \bf{55.9}\% & \bf{93.1\%} & 60.6\% \\\hline
  Ben\cite{benfold2011stable} Head & 45.4\% & 73.8\% & \bf{71.0\%} \\ \hline  
\end {tabular}\label{tab:simi}
\end{minipage}
\caption{\small (Top) Re-detection results for including overlap and co-occurrence terms in the linear relaxed formulation~\eqref{eq:quadlinear}, vs state of the art. The maximum $\Delta t$ considered on the x-axis corresponds to the median length of tracks in the video.
(a) \& (b) Show performance for the MarchingRally and TownCenter sequences
respectively.
In (a), the overlap term significantly improves over NF basic. Interestingly, NF basic outperforms the greedy baseline as opposed to the claim in~\cite{pirsiavash11cvpr} where the difference between the two methods was measured only at $\Delta t=0$ and thus was not visible.
In (b), adding a co-occurrence term to the network flow
formulation also provides a significant improvement over the baselines. (Bottom) Tracking results for TownCenter evaluated in
terms of MOTA measures, compared with results of~\cite{benfold2011stable}.
\vspace{-2mm}} 
\label{fig:aplength}
\end{figure}

For the given $\Delta t$ we collect subtracks from all video intervals $(t, t+\Delta t)$ and sort them according to their confidence.\footnote{The confidence for a subtract in this paper is given by the sum of its constituent detection confidences and correspondence strengths.}
Given the subtrack labels defined above, we evaluate Precision-Recall and Average Precision (AP). High AP values indicate the good performance of the
tracker in the re-detection task. 
On the other hand, common errors such as ID switches and imprecise localization reduce AP values.
Note that in the case of $\Delta t = 0$, our measure reduces to the standard measure for object 
detection. Larger values of $\Delta t$ enable evaluation of re-detection for longer time intervals. 
To compare different methods,
we plot the re-detection AP for different values of $\Delta t$ as illustrated in Figure~\ref{fig:aplength}.

\begin{table*}
  \begin{tabular}{|c|c|c|c|c|c|c|c|c|c|c|c|c|c|}
	\hline  
&        	     & Rcll & Prcn & GT & MT & PT & ML&  FP & FN & IDs & FM & MOTA & MOTP \\ \hline
  \multirow{3}{*}{TUD Stadtmitte} & NF & 67.9 & 72.0 & 10 & 4  & 6  & 0 & 305 & 371&  26 & 26 & 39.3 & 59.5\\ \cline{2-14}
  & NF+pairwise   & 59.6 & \bf{89.9} & 10 & 2  & 8  & 0 &  \bf{77} & 467&  15 & 22 & 51.6 & \bf{61.6}\\ \cline{2-14}
  & Milan~\cite{milan13cvpr}   & \bf{69.1} & 85.6 & 10 & \bf{4}  & \bf{6}  & 0 & 134 & \bf{457}&  15 & \bf{13} & \bf{56.2} & \bf{61.6} \\ \hline \hline
 
  \multirow{3}{*}{PETS S2L1} & NF & 93.7 & 83.4 & 19 & 17 & 2  & 0 & 870 & 293&  64 & 66 & 73.6 & 72.9 \\ \cline{2-14}
				& NF+pairwise   & 92.4 & \bf{94.3} & {19} & \bf{18} & \bf{1}  & 0 & \bf{262} & 354&  56 & 74 & 85.5 & \bf{76.2} \\ \cline{2-14}
  & Milan~\cite{milan13cvpr}   & \bf{96.8} & 94.1 & {19} & 18 & \bf{1}  & 0 & 282 & \bf{148} & \bf{22} & \bf{15} & \bf{90.3} & 74.3 \\ \hline \hline
 
  \multirow{3}{*}{PETS S2L2} & NF   & 47.7 & 87.6 & 43 & 1 & 37  & 5  & 693 & 5383& 291 & 531 & 38.1 & 60.7 \\ \cline{2-14}
				& NF+pairwise   & 60.6 & 88.6 & 43 & 6 & \bf{34}  & 3  & 807 & 4050&  244 & 379 & 50.4 & \bf{60.6}\\ \cline{2-14}
  & Milan~\cite{milan13cvpr}   & \bf{65.1} & \bf{92.4} & 43 & \bf{11} & 31 & \bf{1}  & \bf{549} & \bf{3592} &  \bf{167} & \bf{153} & \bf{58.1} & 59.8 \\ \hline \hline
 
  \multirow{3}{*}{PETS S2L3} & NF &  44.5  &92.2  &44&   9 & 15 & 20 &164& 2428 &121 &189 & 38.0 & 69.3\\ \cline{2-14}
			  & NF+pairwise   &  \bf{45.5}  &91.2  &44&  \bf{12} & 15 & \bf{17} &155& \bf{2125} & 44 &50 & \bf{40.3} & 61.2\\ \cline{2-14}
& Milan~\cite{milan13cvpr}   &  43.0  & \bf{94.2}  &44&   8 & \bf{17} & 19 &\bf{115}& 2493 & \bf{27} & \bf{22}   & 39.8 & \bf{65.0}\\ \hline \hline
 
\multirow{3}{*}{PETS S1L1-2} & NF  & 62.9 &  89.1 & 44 &  18 &  15 &  11 & 295 & 1425 & 289 & 140 & 47.8 & 65.2 \\ \cline{2-14}
& NF+pairwise	 & \bf{68.9} &  92.0 & 44 &  20 &  \bf{16} &  \bf{8} & 230 & \bf{1198} &  35 & 74 & \bf{62.0} & \bf{62.1}  \\ \cline{2-14}
& Milan~\cite{milan13cvpr}   & 64.9 &  \bf{92.4} & 44 &  \bf{21} &  12 &  11 & \bf{169} & 1349 & \bf{22} &  \bf{19} & 60.0 & 61.9 \\ \hline \hline
 
\multirow{3}{*}{PETS S1L2-1} & NF & 31.3 & 87.4 &  42 &  4 & 15 & 23 &208 &3501 &101 &243 & 23.7 & 57.9 \\ \cline{2-14}
& NF+pairwise   & \bf{37.9} & 89.6 &  42 &  \bf{4} & \bf{20} & \bf{18} & 223 & \bf{3141} & 67 & 122 & \bf{32.2} & 55.0 \\ \cline{2-14}
& Milan~\cite{milan13cvpr}   & 30.9 & \bf{98.3} &  42 &  2 & 19 & 21 & \bf{27} &3494 & \bf{42} & \bf{34} & 29.6 & \bf{58.8} \\ \hline

  \end{tabular}
  \caption{Table summarizing results over PETS and TUD sequences. Bold indicates best value for each
  column for each dataset. Abbreviations are as follows GT - ground truth tracks. 
  MT - Mostly tracked. PT - partially tracked. ML - mostly lost. FP - false positives. FN - false negatives.
  IDs - ID swaps. FM - fragmentation. 
  \vspace{-2mm}
  }
  \label{tab:restab}
\end{table*}

\subsubsection{Experimental results}
We compare our algorithm with the state-of-the-art approaches on several video sequences. 
For the Marching\-Rally and TownCenter sequences, the baseline
approaches for comparison are a greedy implementation of the basic min-cost network flow
algorithm with the greedy NMS heuristic from~\cite{pirsiavash11cvpr}, and a network flow (NF) implementation as a linear program. In
all graphs in Figure~\ref{fig:aplength}, the corresponding
results are represented by black (``Greedy + NMS'') and blue (``NF Basic'') curves. We note that we perform a careful grid search over the parameter space for all three algorithms
and show the results corresponding to the best parameters, to make sure the differences observed are not arising from different parameter choices, but rather from limitations of the framework.
On the other hand, we have used only one fixed set of parameter values to produce the results on the different sequences in the PETS and TUD datasets given in Table~\ref{tab:restab}. See~\cite{projectwebpage} for the parameters used and information about the runtime.

In the MarchingRally video sequence, several people are moving
in a crowd in a similar direction. The angle of viewing and the number of people alleviate the issue of clutter, which leads to failure
of tracking algorithms that tend to confuse tracking identities. Our algorithm with
overlap constraints (red curve) outperforms the state of the art by a large margin. Figure \ref{fig:aplength}(a) shows the re-detection accuracy results
with/without the overlap constraints.  Note that the difference in performance between our algorithm and~\cite{pirsiavash11cvpr} grows together with
the re-detection time interval. In fact, for the intervals of $40$ frames or more, our algorithm
outperforms the baseline by over $20\%$ AP. 

The TownCenter sequence is a video with two complementary sets of detections corresponding to
heads and upper bodies.
While head detections are noisy but have high recall, body detections are more precise but are also
prone to more clutter. In such a case, as shown in Figure \ref{fig:aplength}(b) we leverage
body detections to improve noisy head tracks. Again in this case, there is more than $20\%$ 
improvement in AP over the head baseline. Finally, the table in Figure (\ref{fig:aplength}) compares 
our method with a state-of-the-art~\cite{benfold2011stable} algorithm 
in terms of traditional MOTA evaluation measure.
Note that while we compare with a ``greedy'' version
of the overlap term~\cite{pirsiavash11cvpr}, designing a greedy version of the co-occurrence term
is not obvious. 

For the PETS and TUD sequence, we compare the results of our method based on MOTA metrics with those
presented in Milan et al.~\cite{milan13cvpr}. These sequences are challenging for a variety of
reasons. First, there is a crowd of people walking in different directions and criss-crossing each
other, which makes sustained tracking difficult. Second, few full body detections are available
per frame in each video, which makes adding new terms to the objective function difficult. Third,
since people walk side-by-side there is a lot of overlap between detections that belong to two
different persons, hence enforcing the overlap criterion is difficult. 
However, as can be seen in Table~\ref{tab:restab}, our method
generally has comparable MOTA, MOTP and recall scores with~\cite{milan13cvpr}. This shows that
our method is able to address complex scenarios effectively and our cost function is easy to adapt
to general scenarios. Note also that the camera angle in PETS and TUDS are very different from each
other, which means that our algorithm is sufficiently robust to these changes.
Thus, we estimate trajectories better (sum of MT and PT of our method is usually high). 
This also results from the use of both overlap and co-occurrence terms in our
approach, which can take into account head detections as additional information.

\section{Discussion and conclusion}
\label{sec:discussion}
We have presented a generic optimization procedure enabling addition of
quadratic costs to the min-cost network flow tracking methods.
Our method enables modeling of track interactions in a principled way
and provides empirical certificates of small suboptimality.
We have shown practical benefits of our method for two particular
examples of pairwise costs on challenging video sequences.

 Combining different types of pairwise costs into a
 single (linear) cost
 opens up the possibility of tracking complicated motions.
Moreover, while complex cost functions have more tunable parameters, they could be learnt
 from labeled data using structured
 output learning~\cite{lacoste06qap}. This opens up the possibility
 of learning quadratic costs for
 specific  \emph{crowd actions} such as
 panic, street crossing or stampede.

 \section{Acknowledgements}
 This research was supported in part by the projects FluidTracks, EIT ICT Labs, Google research award, ERC grant Activia (no. 307574) and ERC grant LEAP (no. 336845). We thank Patrick Perez for discussions on the multi-target tracking evaluation.

\nocite{andriluka08cvpr}
\nocite{jiang07cvpr}
\nocite{wu06cvpr}

\bibliographystyle{ieee} %
\bibliography{eccvbib}

\clearpage

\appendix

\twocolumn[\centerline{\Large \bf Supplementary Material} \vspace{3em}]

\begin{figure*}[!ht]
  \centering
  \includegraphics[width=1.8in]{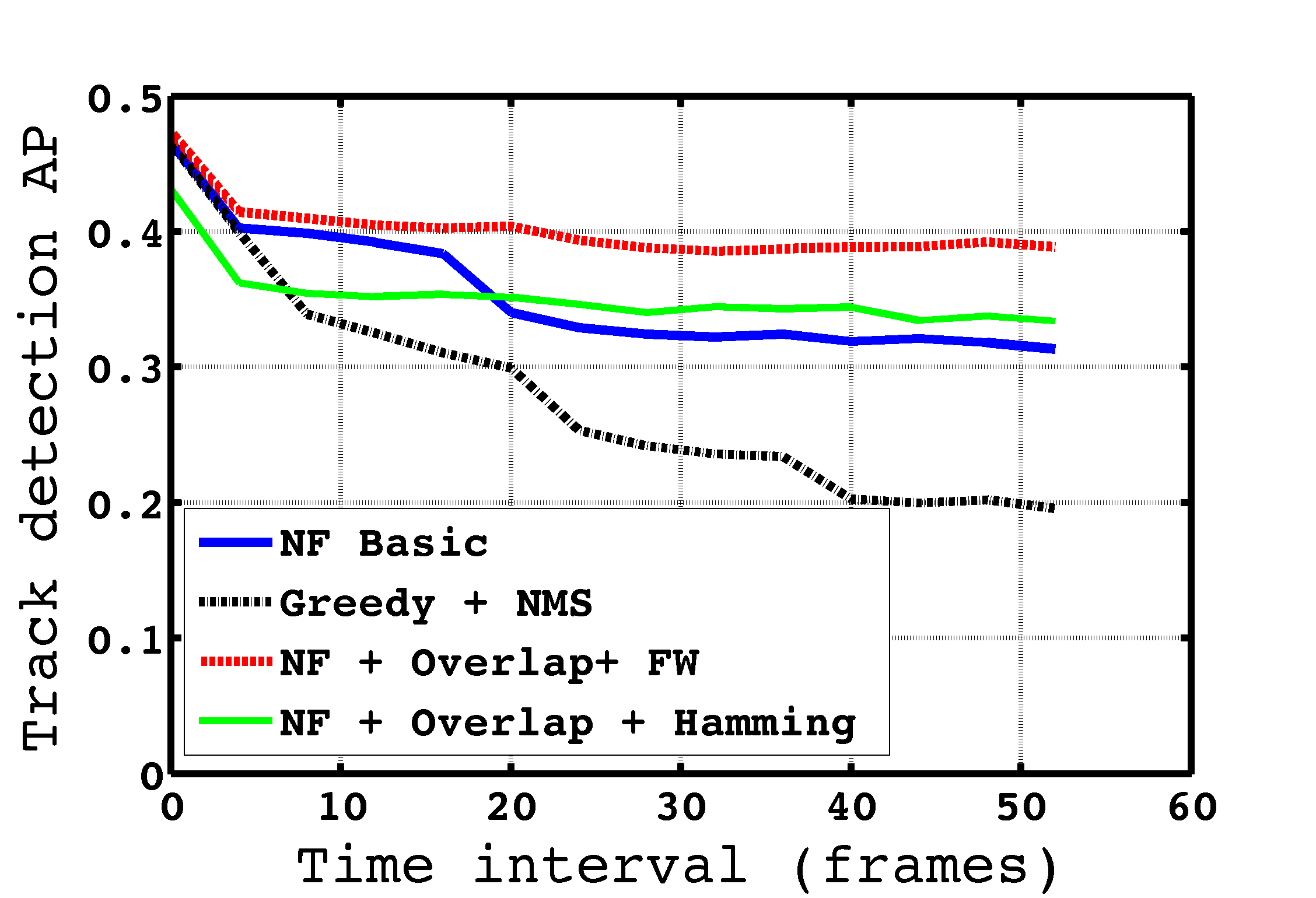}
  \includegraphics[width=1.8in]{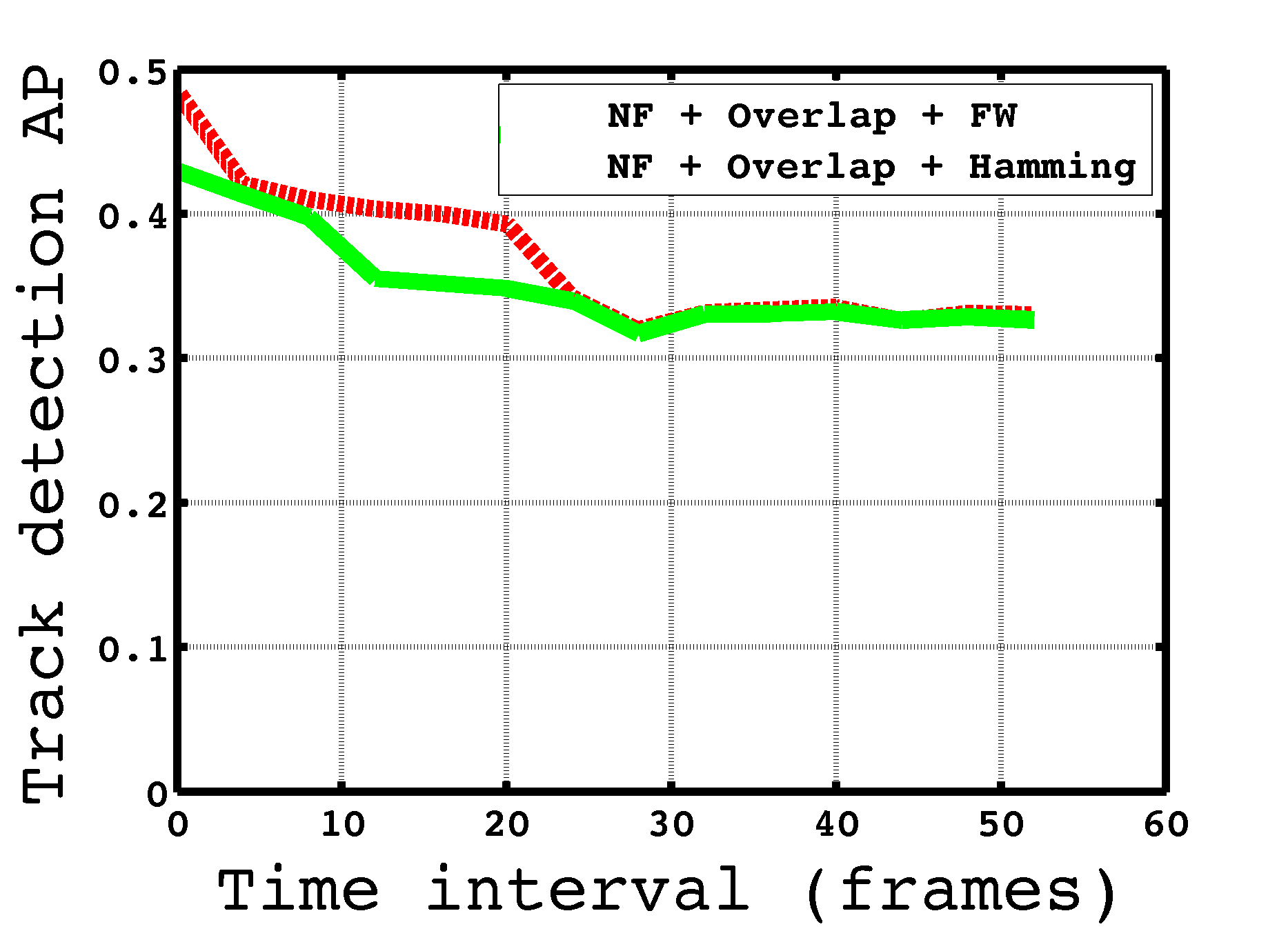}
  \includegraphics[width=1.8in]{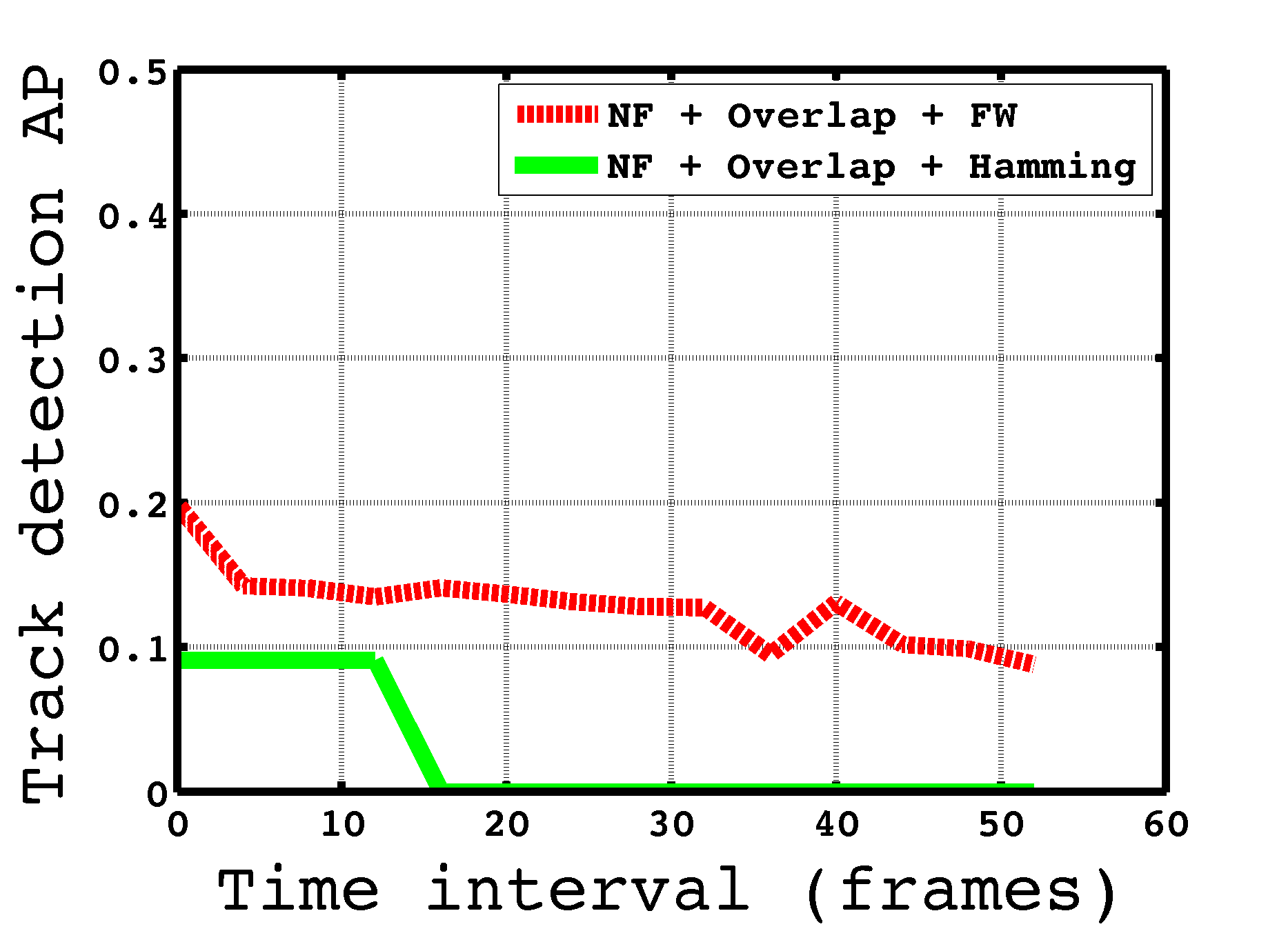} \\
  \includegraphics[width=1.8in]{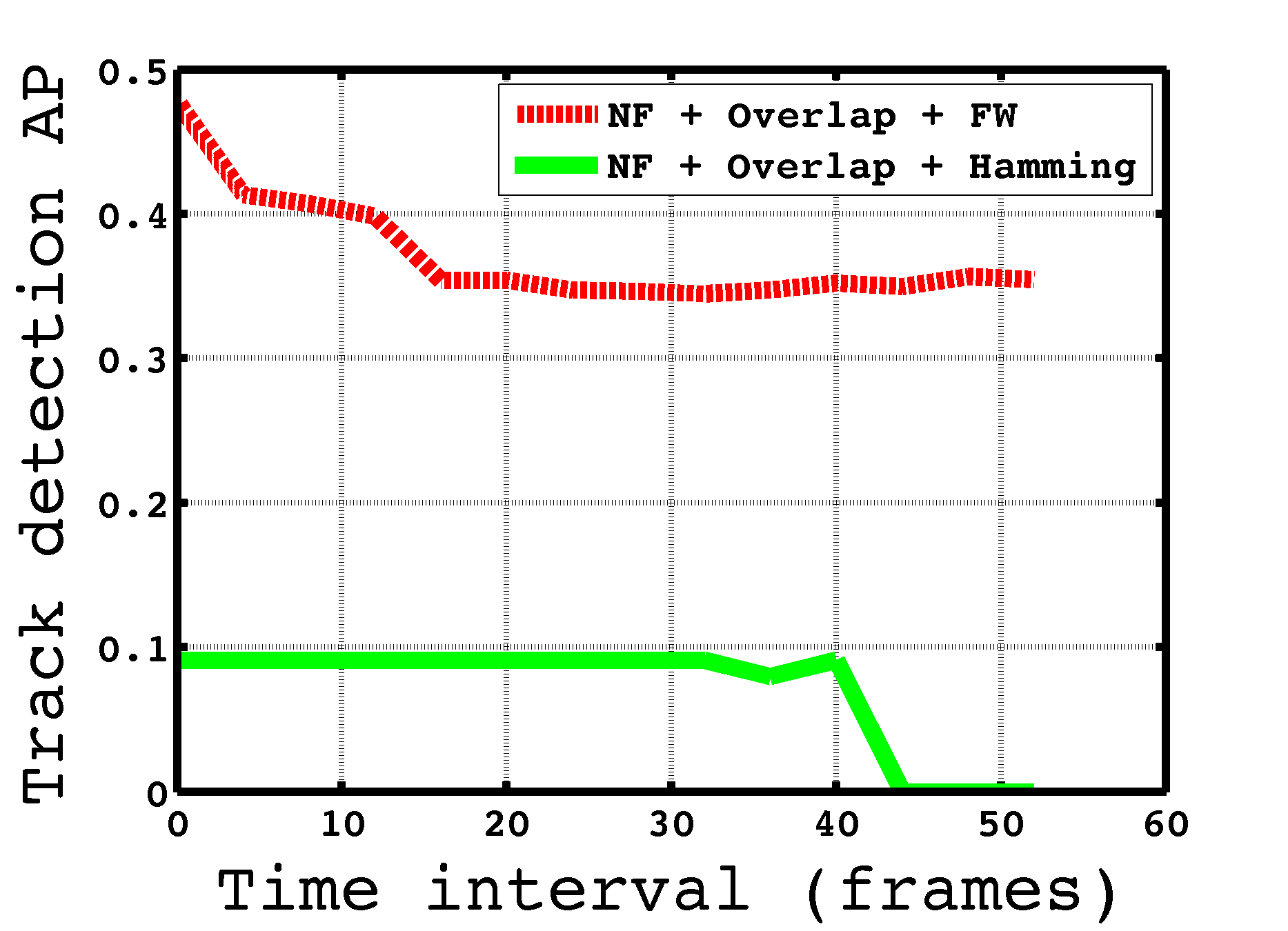}
  \includegraphics[width=1.8in]{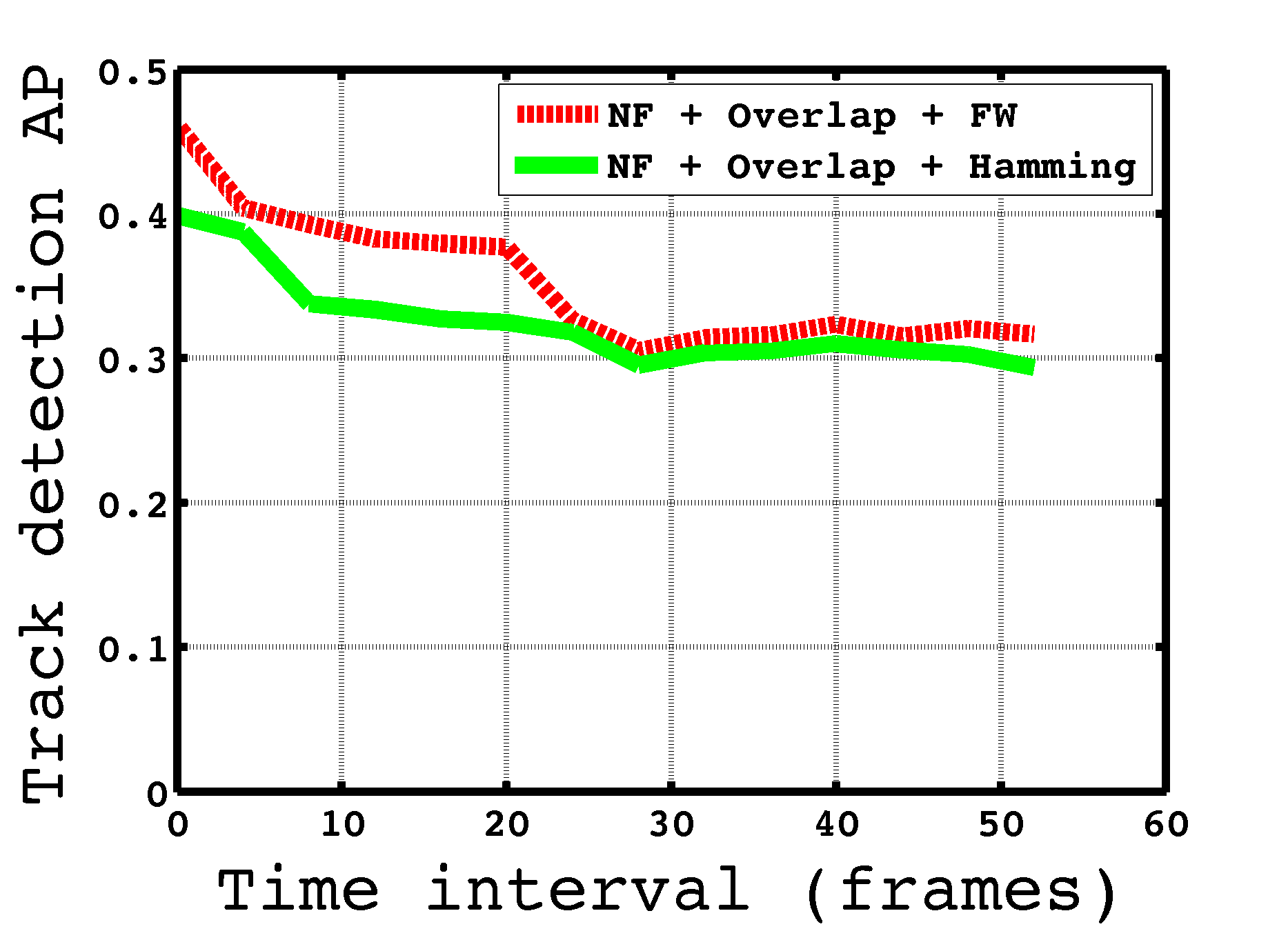}
  \includegraphics[width=1.8in]{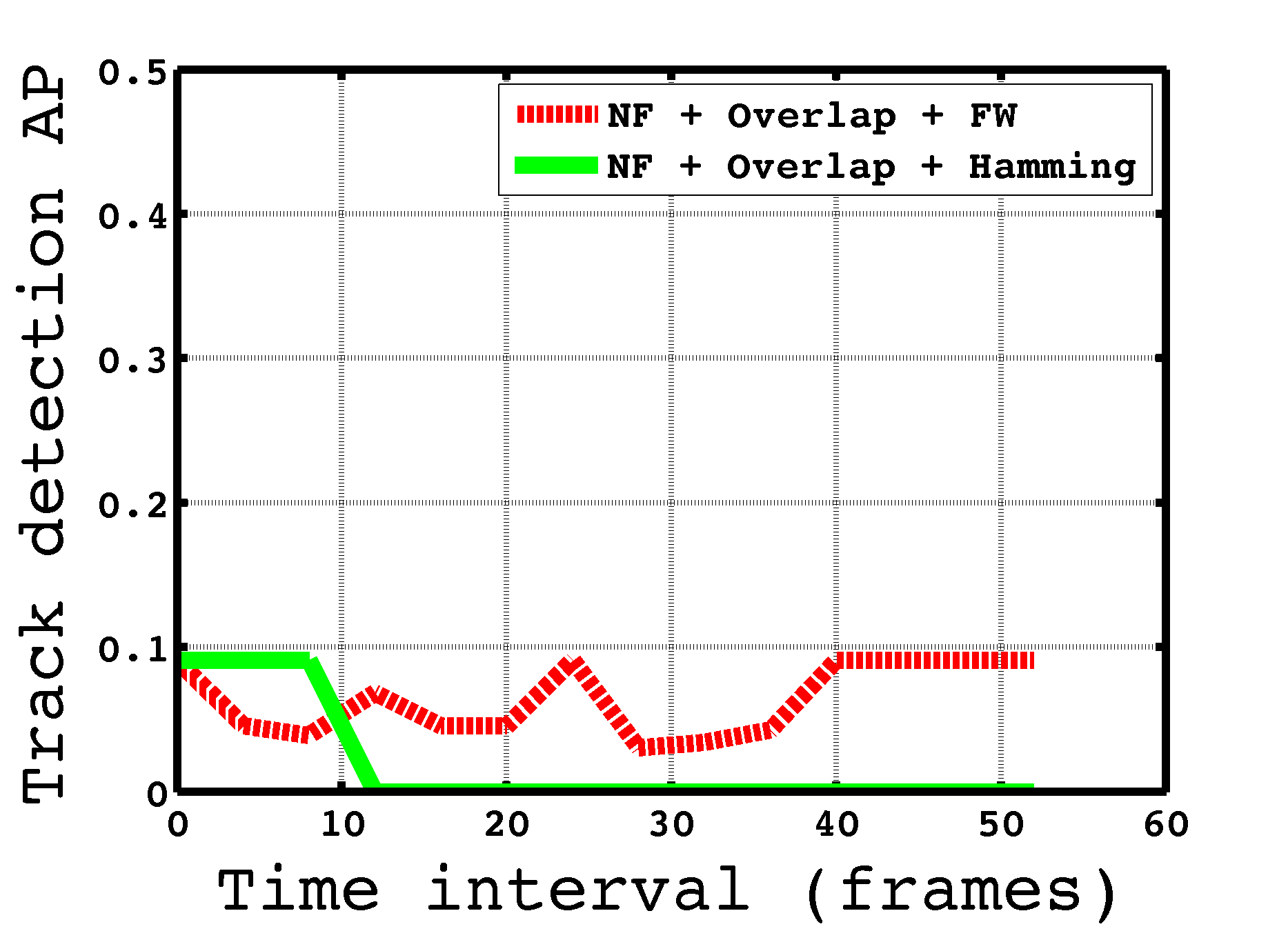}
  \caption{Re-detection accuracy for the cases presented in Table~\ref{table:suboptim} on the MarchingRally sequence. Top row is Case 1 to~3 (varying the overlap penalty); bottom row is Case 4 to~6 (varying the detection confidence weight); Case 1 is Figure~\ref{fig:aplength}(a) in the paper with the additional Hamming rounding curve. Hamming rounding yields systematically worse re-detection accuracy than Frank-Wolfe rounding. }
  \label{fig:hamming}
\end{figure*}        
 
\section{Superiority of Frank-Wolfe rounding heuristic vs. Hamming rounding}
\label{sec:FWbetter}
In Section~\ref{subsubsec:gradient_based_search}, we described two approaches to round the fractional solution $\zb^*$ obtained after optimizing the LP relaxation~\eqref{eq:quadlinear}. ``Rounding'' here meant finding a valid track encoding for prediction, i.e. a $\zb \in \flow_k$ with integer coordinates.
The first approach was to find the vertex (binary vector) in $\flow_k$ with minimal Euclidean distance to $\zb^*$. We called this approach \emph{Hamming rounding} and is standard for problems operating on binary vectors.
We also proposed a novel alternative rounding heuristic called \emph{Frank-Wolfe rounding} which instead minimizes the linear approximation of the quadratic objective~\eqref{eq:quadratic}, and is given by problem~\eqref{eq:gradsearch}.
In our experiments, we observed that Frank-Wolfe rounding yielded solutions with better objective values, as well as better tracking accuracy, than Hamming rounding. We illustrate these observations in this section.

For the MarchingRally experiment (where we only have head detections), we parameterized the objective with two parameters: a multiplicative constant in front of the detection confidences, and the value of the overlap penalty $q_{ij}^{\mathrm{ov}}$ mentioned in~\eqref{eq:overlapterm} (set to a constant).\footnote{We suppose a multiplicative constant of one in front of the correspondence strengths; changing it as well would just amount to multiply the whole objective by a constant, which would not change the solution.} In Table~\ref{table:suboptim}, we compare the suboptimality certificate values for Frank-Wolfe rounding vs. Hamming rounding for 6 different parameter settings on the MarchingRally dataset. More specifically, for each parameter setting, we first obtain the global relaxed solution $\zb^*$ to the LP relaxation~\eqref{eq:quadlinear}, then we either round by Frank-Wolfe rounding or by Hamming rounding and compare their suboptimality certificates. We also compare their re-detection accuracy in Figure~\ref{fig:hamming}, which shows that Frank-Wolfe rounding systematically yields better results than Hamming rounding.

Case 1 in Table~\ref{table:suboptim} is the reference case where we use the best parameter values found by grid search, which were used to produce the results in Figure~\ref{fig:aplength}(a) in the paper. For Case 2 and~3, we vary the overlap penalty weight. 
Case 2 is a very low value for the overlap term encouraging tracks to criss-cross each other, while Case 3 has a very
high overlap weight which means even small amount of overlap is unacceptable. Results for these cases are shown in the first row of Figure~\ref{fig:hamming}. The next three cases vary the weight for 
detection confidence. In particular in Case 6, the presence of negative weight actually
``discourages'' any detections from being picked unless they are connected to edges with extremely
high connection strength. This results in poor performance as shown in Figure~\ref{fig:hamming} but note
that even here, Hamming rounding results are worse than the Frank-Wolfe rounding ones. Also note that worse
suboptimality certificates usually result in worse tracking. 

\begin{table}[!ht]
  \centering
  \begin{tabular}{|c|c|c||c|c|}
	\hline
	& Detection & Overlap & FW & Ham. \\ \hline
	Case1 & 0.1 & 0.0223 & 4.7e-03 & 1.4e-02  \\ \hline
	Case2 & 0.1 & 0.0007 & 8.7e-06 & 9.3e-03\\ \hline
	Case3 & 0.1 & 2.23 &  4.3e-03 & 1.0e-01 \\ \hline
	Case4 & 3.0 & 0.0223 & 9.3e-06& 8.9e-03 \\ \hline
	Case5 & 0.074 & 0.0223 & 3.1e-02 & 1.0e-01\\ \hline
	Case6 & -1.0 & 0.0223 &  1.0e-01& 1.3e-01\\ \hline	
  \end{tabular}
  \caption{Suboptimality certificates for Frank-Wolfe rounding vs. Hamming rounding  on the MarchingRally sequence for different parameter value settings of the objective. The first two columns give the parameter value for the detection confidences and the overlap penalty respectively for each case. The last two columns give the suboptimality certificate for Frank-Wolfe rounding and Hamming rounding (lower is better).}
  \label{table:suboptim}
\end{table}

\section{Video Results}
\label{sec:MOTAvsUS}
\begin{figure*}
  \centering
  \includegraphics[width=5in]{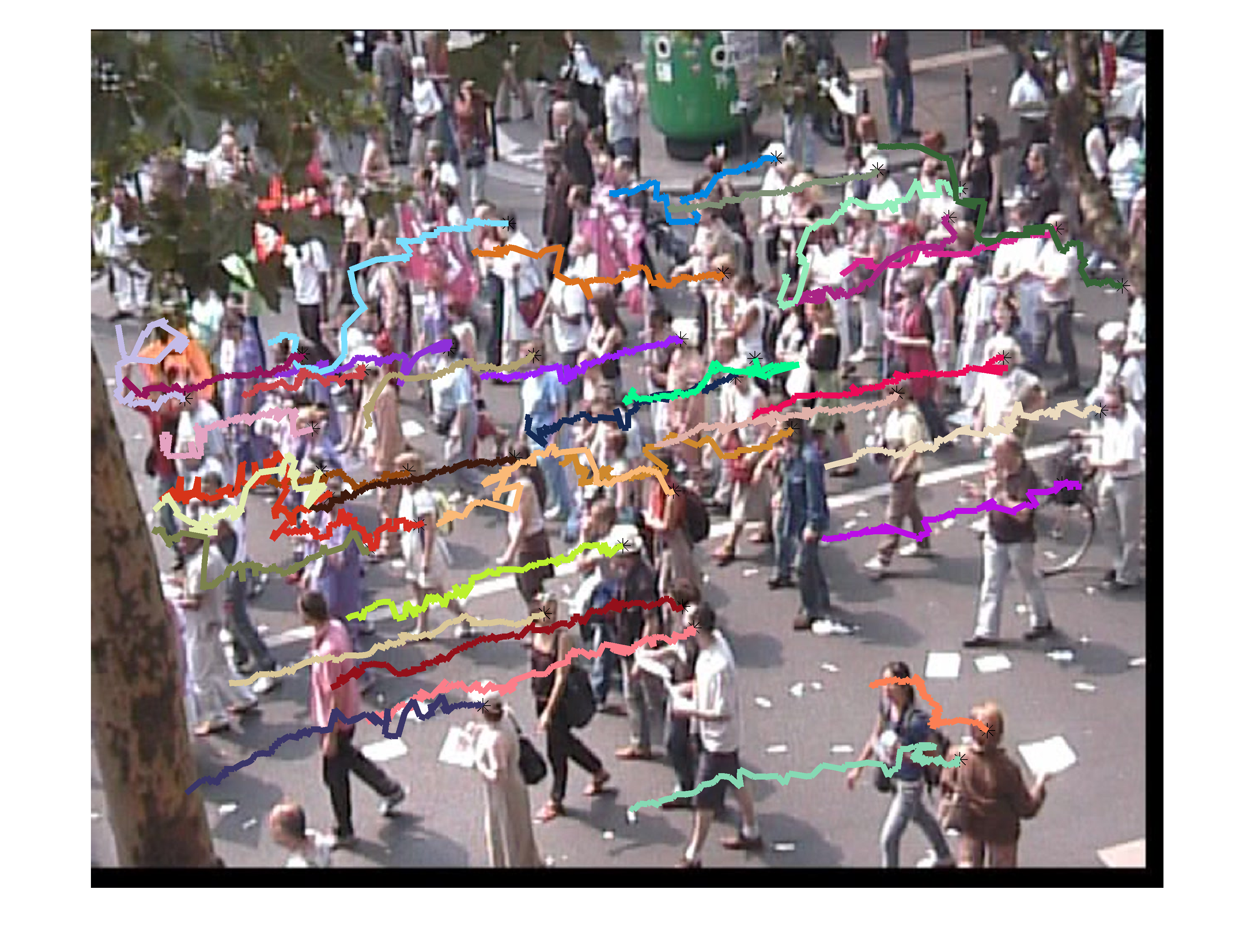} \\
  \includegraphics[width=5in]{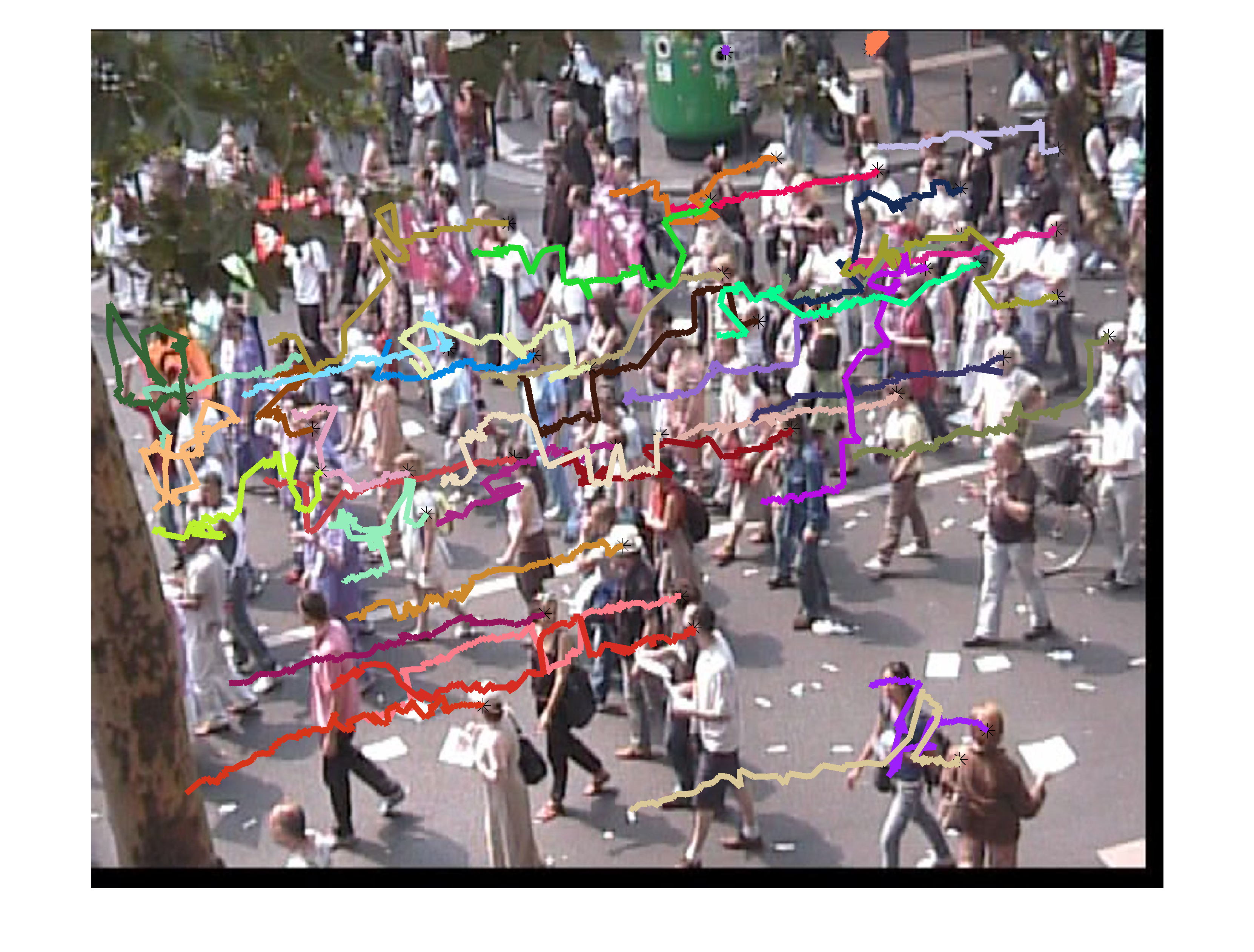}
  \caption{First image of MarchingRally sequence overlaid with tracks. Each track is set to a
  different color. Tracks on top show our result (NF+Overlap), while tracks on the bottom show the result of
  \cite{pirsiavash11cvpr} (Greedy + NMS)}
  \label{fig:tracks}
\end{figure*}

The following images in Figure~\ref{fig:tracks} shows the tracks overlaid on top of the first
frame of the MarchingRally sequence. Each track is shown in a separate color. The output on the
top illustrates our result (NF+Overlap) and the one on the bottom illustrates the results
of~\cite{pirsiavash11cvpr} (Greedy + NMS). Note how in our case one gets non-overlapping tracks while in the case 
of~\cite{pirsiavash11cvpr} there are places where tracks overlap and criss-cross. We
highlight this in videos available from~\cite{projectwebpage}
by drawing cyan colored boxes at places where such ID
swaps happen.
See Figure~\ref{fig:aplength}(a) for the corresponding re-detection curves. For the more classical metrics, the (MOTA, MOTP, IDswap) numbers for NF+Overlap are (27.7\%, 66.5\%, 11) vs. (22.5\%, 66.0\%, 24) for Greedy+NMS.

\section{Runtime and Constraints}
For the PETS and TownCenter dataset, typically we have approximately 10--40 detections per frame.
For PETS data, each detection is connected to a detection in another frame (with a pairwise term)
if they are less than 6 frames apart. On average, each detection is connected to about 10 other detections
for pairwise terms (overlap+CO), which means the number of pairwise terms is linear in the number of unary terms.
For TownCenter data, %
we connect detections
over 30 frames to account for slower motion of people and missing detections, resulting in about 15 pairwise
terms (overlap+CO) per detection on average. While our algorithm runs in about 5--10 seconds on the PETS dataset, it takes about 30--45 minutes on the TownCenter dataset. This difference is due to the larger number of frames in the TownCenter dataset (one order of magnitude greater than for the PETS videos), and also the larger number of pairwise terms per detection on average, resulting in a LP with about 5~million variables in comparison to about 50~thousand for the PETS sequences. 
 
\end{document}